\RequirePackage{fix-cm}
\documentclass[twocolumn]{svjour3}          
\smartqed  
\usepackage{url}
\usepackage{slashbox}
\usepackage{subfigure}
\usepackage{graphicx}
\usepackage{amsmath,amssymb}
\usepackage{multirow}
\usepackage{array}
\usepackage[authoryear,semicolon,sort&compress]{natbib}
\usepackage[ruled,vlined]{algorithm2e}
\usepackage[pagebackref=true,breaklinks=true,letterpaper=true,colorlinks,bookmarks=false,citecolor=blue]{hyperref}
\graphicspath{{figures/}}

\begin{document}
\title{Transferring Object-Scene Convolutional Neural Networks for Event Recognition in Still Images}


\author{Limin Wang \and Zhe Wang \and Yu Qiao \and Luc Van Gool }


\institute{
           Limin Wang \at
              Computer Vision Lab, ETH Zurich, Zurich, Switzerland. \\  \\
	          Shenzhen Institutes of Advanced Technology, Chinese Academy of Sciences, Shenzhen, China. \\
              \email{07wanglimin@gmail.com}
           \and
           Zhe Wang \at
              Shenzhen Institutes of Advanced Technology, Chinese Academy of Sciences, Shenzhen, China. \\
              \email{buptwangzhe2012@gmail.com}
           \and
           Yu Qiao \at
              Shenzhen Institutes of Advanced Technology, Chinese Academy of Sciences, Shenzhen, China. \\
              \email{yu.qiao@siat.ac.cn}
           \and 
	         Luc Van Gool \at
	          Computer Vision Lab, ETH Zurich, Zurich, Switzerland. \\
           \email{vangool@vision.ee.ethz.ch}
}

\date{Received: date / Accepted: date}

\graphicspath{{figures/}}

\maketitle

\begin{abstract}
Event recognition in still images is an intriguing problem and has potential for real applications. This paper addresses the problem of event recognition by proposing a convolutional neural network that exploits knowledge of objects and scenes for event classification (OS2E-CNN).  Intuitively, it stands to reason that there exists a correlation among the concepts of objects, scenes, and events. We empirically demonstrate that the recognition of objects and scenes substantially contributes to the recognition of events. Meanwhile, we propose an iterative selection method to identify a subset of object and scene classes, which help to more efficiently and effectively transfer their deep representations to event recognition. Specifically, we develop three types of transferring techniques: (1) initialization-based transferring, (2) knowledge-based transferring, and (3) data-based transferring. These newly designed transferring techniques exploit multi-task learning frameworks to incorporate extra knowledge from other networks and additional datasets into the training procedure of event CNNs. These multi-task learning frameworks turn out to be effective in reducing the effect of over-fitting and improving the generalization ability of the learned CNNs. With OS2E-CNN, we design a multi-ratio and multi-scale cropping strategy, and propose an end-to-end event recognition pipeline. We perform experiments on three event recognition benchmarks: the ChaLearn Cultural Event Recognition dataset, the Web Image Dataset for Event Recognition (WIDER), and the UIUC Sports Event dataset. The experimental results show that our proposed algorithm successfully adapts object and scene representations towards the event dataset and that it achieves the current state-of-the-art performance on these challenging datasets.
\end{abstract}


\section{Introduction}
\label{sec:intro}

Image classification is a fundamental and challenging problem in computer vision and many research efforts have been devoted to this topic during the past years \citep{KrizhevskySH12,IoffeS15,EveringhamGWWZ10,SimonyanZ14a,ZhouLXTO14,HeZRS15}. The majority of these contributions focus on the problem of object recognition and scene recognition, partially due to the simplicity of object and scene concepts and the availability of large-scale datasets (\emph{e.g.} ImageNet \citep{DengDSLL009}, Places \citep{ZhouLXTO14}). On the other hand, event recognition \citep{WangWDQ15,Salvado,Park,XiongZLT15,LiF07} in static images also is important for semantic image understanding. Being able to selectively retrieve event images helps us keep nice memories of particular episodes of our lives or to more effectively find relevant illustrations, helps to locate where images were taken or to analyze people's culture and so on. 

\begin{figure}
\includegraphics[width=\linewidth]{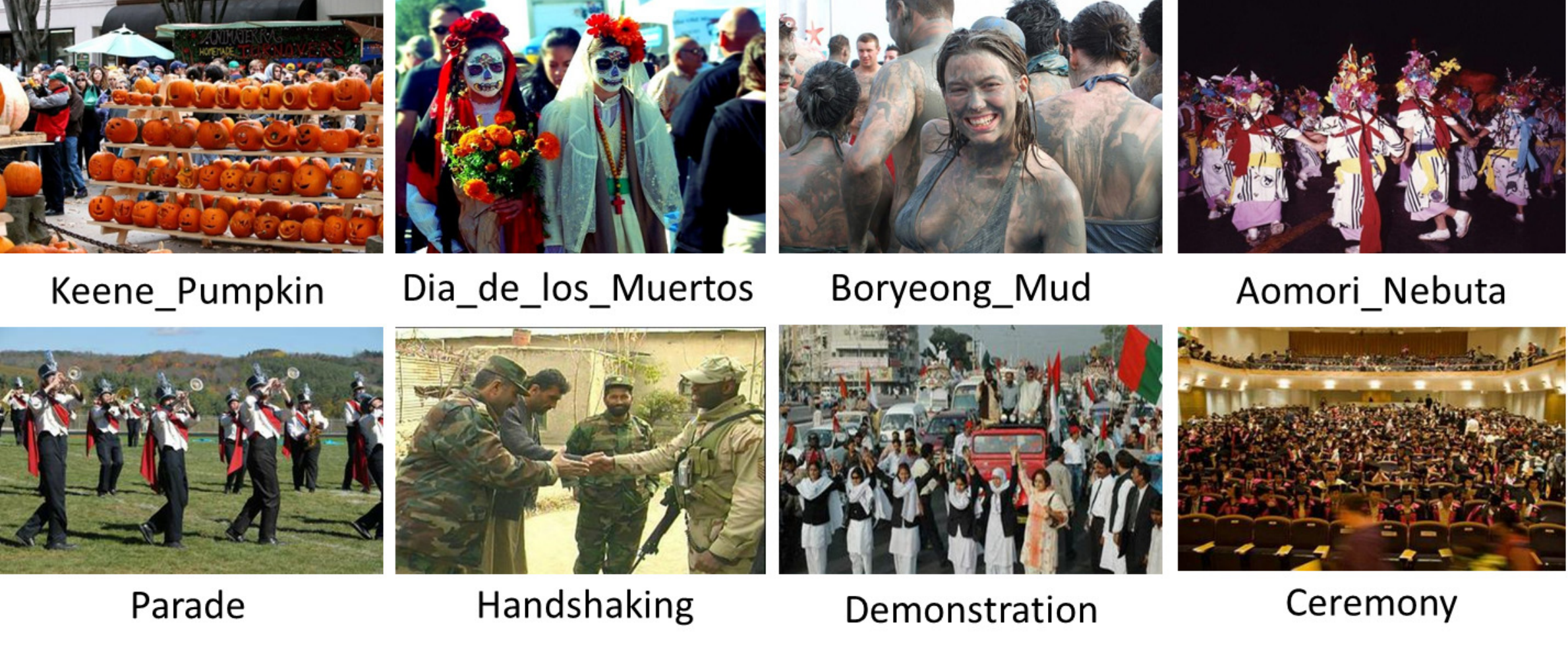}
\caption{Examples of event images from the ChaLearn Cultural Event Recognition dataset (top row) and the Web Image Dataset for  Event Recognition (WIDER) (bottom row).}
\label{fig: dataset}
\end{figure}

In general, an event captures the complex behavior of a group of people, interacting with multiple objects, and taking place in a specific environment.  As illustrated in Figure~\ref{fig: dataset}, the characterization of the concept `event' is relatively complicated compared with the concepts of objects and scenes. Images from the same event category may vary even more in visual appearance and structure. Multiple high-level semantic cues, such as interacting objects, scene context, human poses, and garments, can provide useful cues for event understanding and should be exploited.

Recently, Convolutional Neural Networks (CNNs) \citep{lecun-98} have delivered great successes in large-scale image classification, in particular for object recognition \citep{KrizhevskySH12} and scene recognition \citep{ZhouLXTO14}. Large-scale image datasets \citep{DengDSLL009,ZhouLXTO14} (more than 1 million images) with supervised labels have proven of great importance in this. However, for event recognition, the current public datasets are relatively small and could be easily over-fitted by deep models. Moreover, the inherent complexity of the concept of `event' increases the difficulty of training an event CNN from raw images. Therefore, transferring deep models successfully trained on other datasets to the case of event recognition comes out to be a practical approach. Specifically, in this paper, we aim to study \emph{why the deep representations learned from object and scene datasets are helpful for event recognition} and \emph{how to effectively transfer these deep representations for more accurate event recognition}. In this vein, we make three contributions to event recognition in still images that are described in the next paragraphs.

First, we empirically investigate the correlation among object, scene, and event recognition. As part of this effort, we identify a subset of object and scene categories which are discriminative for event recognition. Essentially, events are complex social phenomena composed of multiple people in the presence of relevant objects and within a specific type of scene. Therefore, an event class will co-occur with certain objects and scenes, while having low correlation with others. Hence, we explore these inherent relationships between objects, scenes, and events in a quantitative way, and provide some evidence that it is a good choice to transfer object and scene deep models to event recognition. Specifically, thanks to the large-scale object and scene datasets, we utilize these pre-trained deep object and scene models to test the images of different event classes. The resulting prediction scores enable us to study the joint distribution of objects, scenes, and events, and analyze the correlation of these 3 concepts. Furthermore, based on these empirical observations, we propose a method to select a subset of object and scene classes, that have a strong discriminative capacity to recognize event classes. This subset of object and scene categories will allow us to better transfer the deep object and scene models to the target of event recognition.

Second, we propose new frameworks to transfer these deep representations learned on the ImageNet dataset and the Places205 dataset to event recognition. Due to the smaller size of the training datasets, it would be extremely difficult to train a deep network from scratch for event recognition. Therefore, we propose three different transferring techniques: (1) initialization-based transferring, (2) knowledge-based transferring, and (3) data-based transferring. In the scenario of initialization-based transferring, we simply copy the weights of the convolutional layers from the object or scene models to those of event models and randomly initialize the fully-connected layers. After this, we carefully fine-tune the whole networks with event supervision information from the target dataset. In the latter two scenarios, we design a multi-task learning framework to regularize the fine-tuning procedure of the event networks and improve their generalization ability. For instance, in knowledge-based transfer, we utilize the object and scene models to guide the fine-tuning of event networks, and we leverage the soft outputs of object and scene CNNs as supervision signals to assist with the training of event networks. In the case of data-based transfer, we exploit the large-scale object and scene datasets to regularize the training process of event networks. We propose to jointly train the event networks with object and scene networks under a weight-sharing scheme for common convolutional layers. 

Finally, based on these learned event CNNs, coined ``OS2E-CNNs'', we propose a simple yet effective event recognition pipeline. Specifically, we design an end-to-end event recognition method without utilizing explicit encoding methods or learning additional classifiers. First, we design a multi-ratio and multi-scale image cropping strategy and transform each image into a set of regions. This multi-ratio and multi-scale strategy is able to deal with scale variations and aspect ratio differences in event images. Then, these image regions are fed into OS2E-CNNs for event recognition. The two-stream OS2E-CNNs can incorporate visual cues for both scenes and objects. Finally, the prediction scores from multiple image regions are averaged to obtain the final recognition result. Based on our proposed event recognition pipeline, we perform experiments on three event recognition datasets: the ChaLearn Cultural Event Recognition dataset \citep{ICCV_LAP}, the Web Image Dataset for Event Recognition (WIDER) \citep{XiongZLT15}, and the UIUC Sports Event dataset \citep{LiF07}.  The experimental results demonstrate that our event recognition method obtains the state-of-the-art performance on all these challenging datasets.

The remainder of this paper is organized as follows. In Section \ref{sec:rw}, we review work related to ours. Then, we present an empirical study of the correlation of objects and scenes with events in Section \ref{sec:corr}. Section \ref{sec:transfer} gives the description of our proposed transferring techniques of adapting object and scene CNNs for event recognition. We propose a simple yet effective event recognition pipeline with our learned OS2E-CNNs in Section \ref{sec:recognition}. The experimental evaluation and exploration is described in Section \ref{sec:exp}. Finally, we discuss our method and offer some conclusions in Section \ref{sec:conclusion}.


\section{Related Works}
\label{sec:rw}

In this section, we briefly review previous work that is related to ours, and highlight the differences.

\textbf{Event recognition.} The analysis of human actions and events is an active research area and much of the prior art has focused on video data \citep{DuanXTL12,BhattacharyaKSS14,WangKSL13,WangOVS15,WangQT15a,WangQT15b,SimonyanZ14,TangYLK13}. With their additional temporal dimension, videos provide motion information that is useful for human behavior understanding. Therefore, those methods heavily rely on the temporal dynamics and can not be readily adapted to the image domain for event recognition. 

For still images, action recognition \citep{YaoJKLGF11,YaoF10,DesaiR12,DelaitreSL11,GkioxariGM15} tends to receive more attention than event recognition. The major difference between actions and events is that the concept of action usually focuses on describing the behavior of a single person, while event recognition usually tries to identify the activity of a group of people, taking place in a specific environment. Among the work on event recognition in still images, \cite{LiF07} proposed a coupled LDA framework to jointly infer the category of event, object, and scene. This method coupled two LDA models and is difficult to scale up to large-scale datasets. \cite{XiongZLT15} designed a deep CNN architecture by fusing the responses of multiple deep channels for objects, faces, and people, and performed event recognition in a end-to-end manner.  Recently, at the ChaLearn Looking at People (LAP) challenge \citep{CVPR_LAP,ICCV_LAP}, several deep learning based methods for the task of cultural event recognition were presented~\citep{Wei15,Liu15,Rothe15,WangWDQ15,WangWGQ15}. \cite{Liu15} proposed to use selective search to generate a set of proposals and to exploit a feature hierarchy to represent these proposals for event recognition. \cite{Wei15} designed a ensemble framework to incorporate the spatial structure into deep representations within a deep spatial pyramid framework \citep{GaoWWL15}. \cite{Rothe15} developed a deep linear discriminative retrieval approach for event recognition by extracting features from four layers of CNNs.  

This paper is based on our previous challenge papers \citep{WangWGQ15,WangWDQ15}, where we were the first to propose the object-scene convolutional neural networks (OS-CNNs) to transfer the deep representations of object and scene models to the task of event recognition. This architecture was widely adopted by other participants~\citep{Wei15,Liu15,Rothe15} at the ICCV ChaLearn LAP challenge~\citep{ICCV_LAP}. We substantially extend our previous work by empirically investigating the correlation of objects, scenes, and events, identifying a small subset of discriminative object and scene classes, and proposing new transferring techniques to improve the generalization capacity of learned event models. 

\textbf{Objects, scenes, and events.}  Events and actions have been shown to correlate with relevant objects and scenes. Examples of papers on the relations between objects, scenes, and actions are \cite{LiF07,GuptaD07,MarszalekLS09,VuOLOS14,Jain15,SrikanthaG14}. \cite{LiF07} proposed a unified framework to jointly perform object, scene, and event recognition through a generative graphical model. \cite{MarszalekLS09} used scene information as contextual cues to improve action recognition performance in videos. \cite{VuOLOS14} also used scene information to predict actions, but for still images. On the other hand, \cite{GuptaD07} and \cite{SrikanthaG14} used actions and events as context helping to disambiguate similarly looking objects and to improve the performance of object recognition. Recently, \cite{Jain15} integrated object CNNs to action recognition. 

These previous results on the correlation of objects, scenes, and events almost come with hand-crafted features and the number of object and scene categories is quite limited. Our work differs in two ways: (1) we scale up the number of objects and scenes by using deep models pre-trained on the ImageNet dataset and Places205 dataset. These large-scale object and scene datasets allow us to explore the correlations of objects and scenes with events in a more realistic scenario. (2) we also propose a new transfer technique to adapt object and scene representations for the task of event recognition, and our transferring technique is based on our empirical study of the aforementioned correlations.

\textbf{Transfer learning.} Many approaches have been proposed in recent years to solve the visual domain adaption problem \citep{FernandoHST13,GongGS14,KulisSD11}, also called the visual dataset bias problem \citep{TorralbaE11}. These methods recognized that there is a shift in the distribution of the source and target data representations, and they usually tried to learn a feature space transformation to align the source and target representations. Recently, supervised CNN representations have been shown to be effective for a variety of visual recognition tasks \citep{Girshick14,Oquab14,ChatfieldSVZ14,Razavian14,Azizpour15}. \cite{Razavian14} proposed to treat CNNs as generic feature extractors, yielding an astounding baseline for many visual tasks. \cite{Girshick14} designed a region-based object detection pipeline and transferred classification models to the detection task. \cite{Oquab14} proposed a transferring framework to adapt a representation learned from the ImageNet dataset for various tasks on the Pascal VOC dataset \citep{EveringhamGWWZ10}. \cite{ChatfieldSVZ14} comprehensively studied three types of models pre-trained on the ImageNet dataset and transferred these representations on the Pascal VOC dataset. \cite{Azizpour15} empirically analyzed different factors of transferability for a generic CNN representation on a variety of tasks. \cite{Tzeng15} recently proposed to jointly learn deep models between source and target domains, but both domains shared the same task.

Our proposed transferring techniques aim to tackle the problem of domain adaption and target adaption simultaneously, where both the distributions of input images and final targets are different from those of the source task. We propose a multi-task learning framework, which tries to utilize extra knowledge or data to guide the fine-tuning procedure on the target dataset. Therefore, our proposed transferring methods can help to reduce the effect of over-fitting and improve the generalization ability of the learned models. Similar soft targets have been exploited for other tasks, such as model compression~\citep{HintonVD15} and training deeper student networks in multiple stages \citep{RomeroBKCGB14}, but our goal is completely different.


\section{An Empirical Study}
\label{sec:corr}

In this section, we  conduct an empirical study on the following two questions: (1) \emph{how to quantitatively evaluate the correlation of events with common object and scene classes} and (2) \emph{how to select a subset of discriminative object and scene categories for event recognition}.

\subsection{Evaluating object and scene responses}

\begin{figure*}[t]
\includegraphics[width=\textwidth]{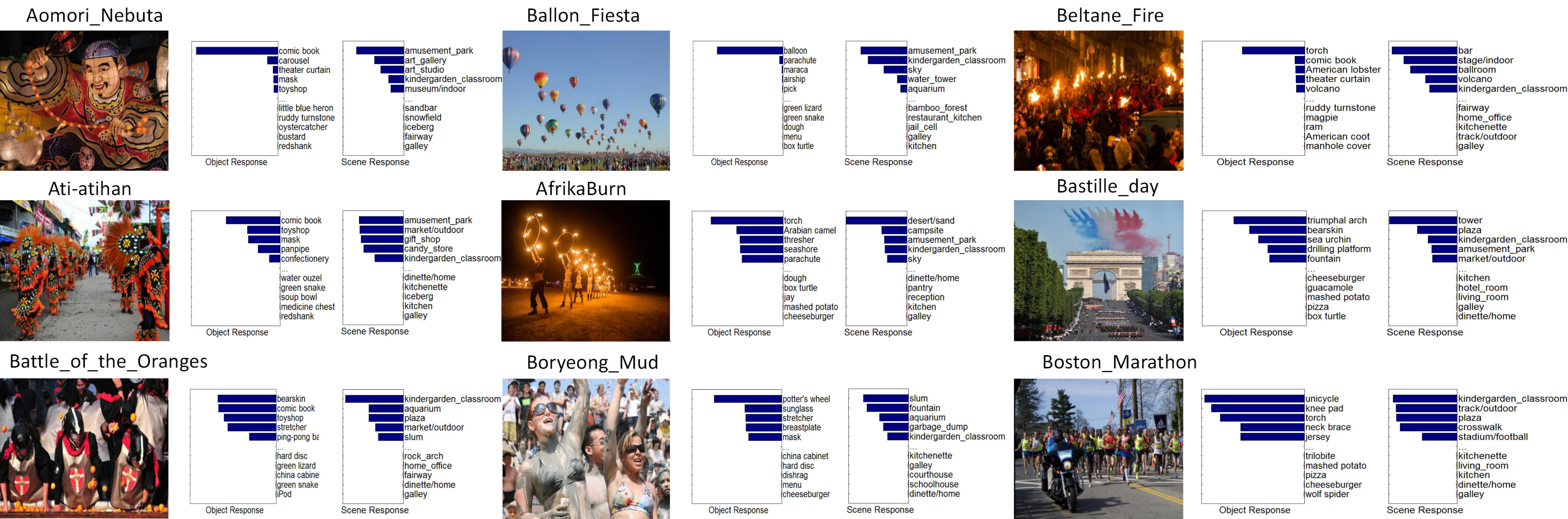}
\caption{ChaLearn Cultural Event categories with corresponding histograms of object responses (\emph{i.e.} $p(o|e)$) and scene responses (\emph{i.e.} $p(s|e)$). Top row: event categories with relatively low-entropy object responses. Middle row: event categories with relatively low-entropy scene responses. Bottom row: event categories with high-entropy object and scene responses. Best viewed in color. } 
\label{fig:os_response}
\end{figure*}

In order to empirically study the correlation between the presence of objects / scenes and event classes, we choose CNNs pre-trained on the large-scale ImageNet dataset \citep{DengDSLL009} and Places205 dataset \citep{ZhouLXTO14} as the generic object and scene detectors, respectively. The two datasets are currently the largest object recognition and scene recognition datasets, where ImageNet contains 1,000 object classes and Places205 includes 205 scene classes. These two datasets cover almost all common object and scene classes, that possibly correlate with the event classes. We now first introduce details on the training of these object and scene CNNs. Then, we describe how to calculate the object and scene responses for each image.

\textbf{CNN models.} We choose the architecture of inception with Batch Normalization (inception-bn) \citep{IoffeS15}, due to its balance between accuracy and efficiency. This network starts with 2 convolutional and max pooling layers, subsequently has 10 inception layers, and ends with a global average pooling layer and a fully connected layer. For both object and scene CNN models, we use the same training policy. Specifically, we use a multi-GPU parallel version \citep{WangXW015} of the Caffe toolbox \citep{JiaSDKLGGD14}, which is publicly available online  \footnote{\url{https://github.com/yjxiong/caffe}}. These networks are learned using the mini-batch stochastic gradient decent algorithm, where the batch size is set to 256 and momentum to 0.9. The learning rate decreases exponentially and reduces to 10\% every 200,000 iterations. The training procedure stops at 750,000 iterations. Concerning data augmentation, we use the normal techniques such as fix cropping, scale jittering, and horizontal flipping \citep{WangXW015}. The training images are resized to $256 \times 256$ and these cropped regions are resized to $224 \times 224$ . These regions have the same labels as the original images and are fed into CNNs for training. Our inception-bn models achieve a top-5 error of $7.9\%$ on the ImageNet validation dataset and $11.8\%$ on the Places205 validation dataset with a single crop, which are comparable to the performance reported in \cite{IoffeS15} on the ImageNet validation dataset.

\textbf{Object and scene responses.} After the training of the object and scene CNN models, we treat them as generic object and scene detectors. More concretely, we choose the ChaLearn Culture Event Recognition dataset \citep{ICCV_LAP}  for our empirical investigation. It should be noted that our exploration and analysis method can also be applied to other event datasets. We use these object and scene CNN models to scan over the training dataset and compute the likelihood of the existence of certain object and scene classes for each image.  Specifically, we first resize each image into three different scales of $256 \times 256$, $384 \times 384$, and $512 \times 512$ to handle scale variations. Then, we crop $224 \times 224$ regions from these resized images by a $3 \times 3$ grid. Finally, each crop is fed into the CNN models to obtain score vectors $S^o$ and $S^s$ to describe the distribution of object and scene classes, respectively. These scores of multiple crops from a single image are averaged to yield the image-level object and scene distribution $\Phi^o$ and $\Phi^s$.

\subsection{Exploring the responses of objects and scenes}
After having calculated the object and scene responses for the training images, we are ready to quantitatively analyze the correlation among these three concepts: objects, scenes, and events. Here we try to answer the question of \emph{whether and how different event classes tend to co-exist with distinctive sets of objects or scenes}. 
 
More specifically, given a set of event training images $\{\mathbf{I}_i, y_i, \Phi^o(\mathbf{I}_i), \Phi^s(\mathbf{I}_i)\}_{i=1}^N$, where $\mathbf{I}_i$ denotes the image, $y_i$ is its event label, and $\Phi^o(\mathbf{I}_i)$ and $\Phi^s(\mathbf{I}_i)$ are the scores of object and scene responses calculated as described in the previous subsection, we estimate the conditional distribution $p(o|e)$ and $p(s|e)$ as follows:
\begin{equation}
\begin{split}
p(o|e) = \frac{1}{N_e} \sum_{i: y_i = e} \Phi^o(\mathbf{I}_i)[o], \\
p(s|e) = \frac{1}{N_e} \sum_{i: y_i = e} \Phi^s(\mathbf{I}_i)[s],
\end{split}
\label{equ:likelihood}
\end{equation}
where $p(o|e)$ and $p(s|e)$ are the conditional distribution of object class $o$ and scene class $s$ given event class $e$ respectively, $\Phi[j]$ represents the $j^{th}$ element of $\Phi$, and $N_e$ is the number of images belonging to the $e^{th}$ event class (\emph{i.e.} $N_e = \sum_{i=1}^N \mathbb{I}(y_i=e) $). We take the average of images from the same event class to estimate the conditional distribution of objects and scenes given a specific event class. Meanwhile, we estimate the prior probability of event $e$ as follows:
\begin{equation} 
p(e) = \frac{N_e}{N},
\label{equ:prior}
\end{equation} 
where $N$ is the total number of training images.

\begin{figure*}
\includegraphics[width=\linewidth,height=0.32\linewidth]{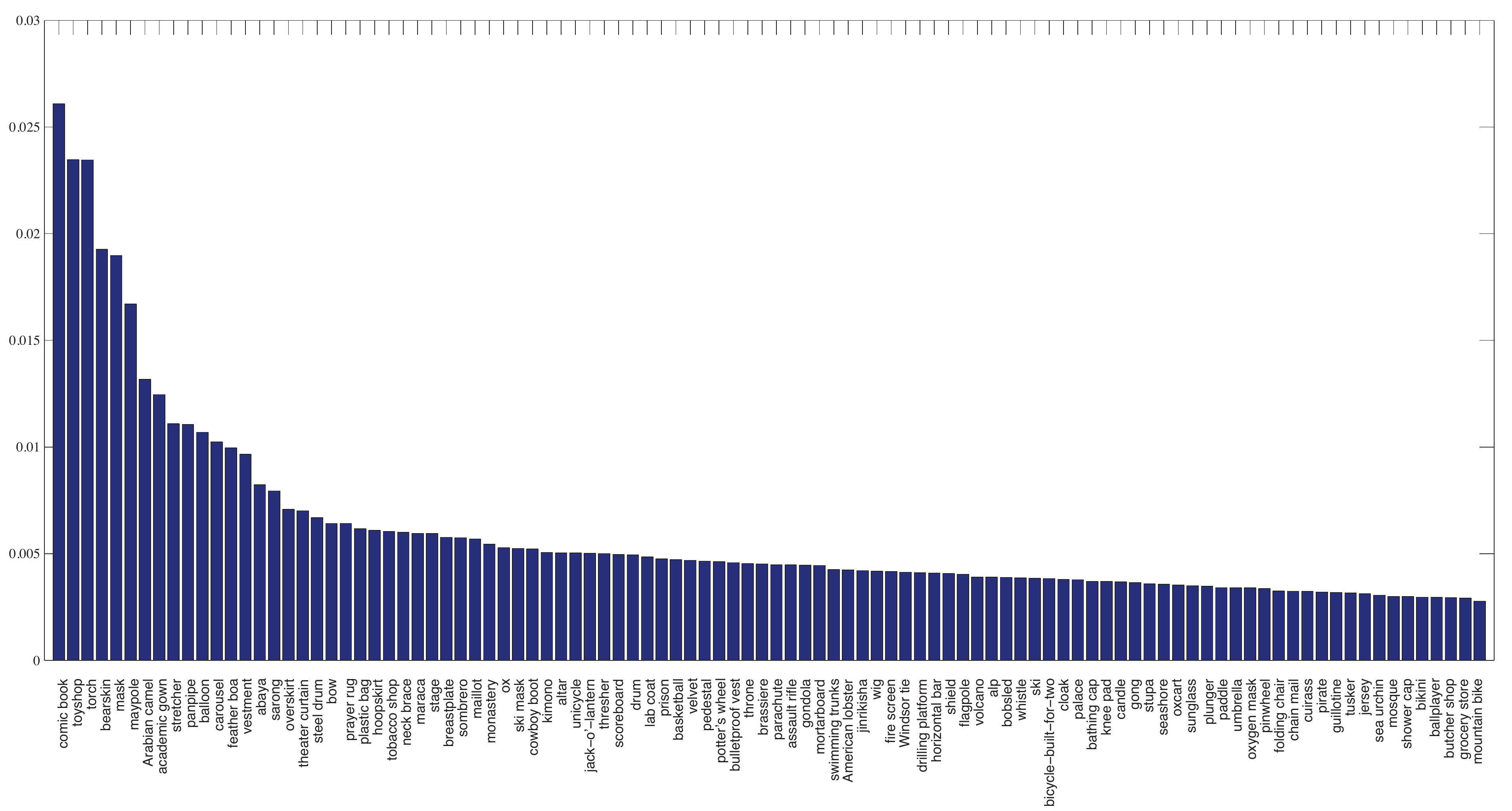}
\includegraphics[width=\linewidth,height=0.35\linewidth]{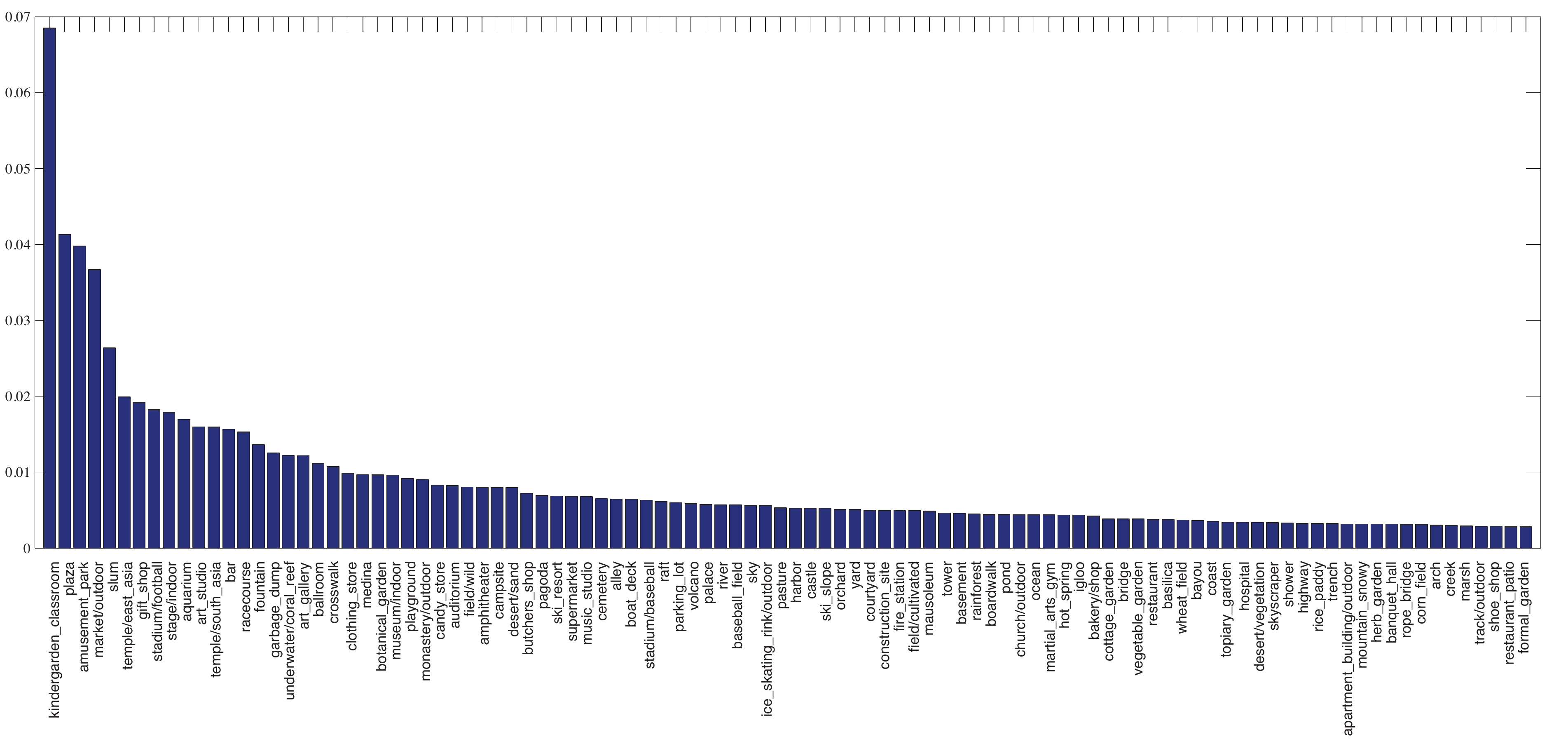}
\caption{The overall object responses (\emph{i.e.} p(o)) and scene responses (\emph{i.e.} p(s)) on the ChaLearn Cultural Event Recognition dataset. We notice that the distributions of object responses and scene responses yielded a ``long tail'', which means the most probability goes to a small portion of the object and scene classes. Best viewed in color.}
\label{fig:os_distribution}
\end{figure*}

\begin{table}
\caption{Event recognition results using the likelihood that objects and scenes are present, for the ChaLearn Cultural Event Recognition dataset under the validation settings. Objects and scenes can provide important visual cues for event recognition, which achieves a relatively high recognition performance. The combination of objects and scenes can further boost the recognition performance.}
\label{tbl:result1}
\centering
\begin{tabular}{|l|c|c|c|}
\hline
Responses & objects & scenes & combination \\ 
\hline
Performance (mAP) & 70.2\% & 61.5\% & 72.1\% \\
\hline
\end{tabular}
\end{table}

To investigate the correlation between objects or scenes and events, we visualize several examples of conditional probabilities $p(o|e)$ and $p(s|e)$ in Figure \ref{fig:os_response}. We notice that some event classes show strong responses to specific object or scene classes. For example, in the event of \texttt{aomori nebuta}, the object class \texttt{comic book} has the highest response. In the event of \texttt{beltane fire}, there is a strong preference for the object class \texttt{torch}.  For scene responses, the event of \texttt{afrika burn} yields a high response for the \texttt{desert} scene class and the event of \texttt{bastille day} for the \texttt{tower} scene class. In such cases, the corresponding object or scene classes may act as a strong visual attribute that can be exploited for event recognition. On the other hand, some event classes, such as \texttt{battle of the orange} and \texttt{boston marathon}, may have strong response scores for multiple object and scene classes simultaneously. In these situations, the co-occurrence of several object and scene classes may contribute to the prediction of event classes. From these examples in Figure \ref{fig:os_response} we conclude that each event class may have a strong response to a specific object and scene class, or a small subset of object and scene classes.

To further evaluate the contribution of objects and scenes to event recognition, we conduct experiments by using the object response $\Phi^o(\mathbf{I})$ and the scene response $\Phi^s(\mathbf{I})$ as image representations. In this experiment, we train one-vs-all SVMs as classifiers. In order to make the training of the SVMs more stable, we first normalize these responses with $\ell_2$-norm \citep{VedaldiZ12}.  The numerical results are reported in Table \ref{tbl:result1}. We find that the object responses achieve a mAP of $70.2\%$ and scene responses obtain a mAP of $61.5\%$. The object responses outperform those for scenes for the task of event recognition. The combination of both responses further improves the performance to $72.1\%$. We can conclude that the context information provided by the present objects and scene is a useful cue for event recognition. This recognition performance inspires us to further transfer the object and scene representations to event recognition.

In summary, we empirically found that the occurrence of some event classes strongly correlates with the presence of specific object and scene classes. Thus, utilizing the object and scene likelihood as semantic features, could enable event CNNs to start from a meaningful initialization and converge to a stationary solution during event CNN training. We will explore how to transfer these object and scene CNN models to the task of event recognition in Section \ref{sec:transfer}.

\subsection{Selecting objects and scenes}
\label{sec:selection}

A natural question arises \emph{whether all common object and scene classes are necessary to accomplish the task of event classification}. In this subsection, we aim to find a subset that is most discriminative for distinguishing the event classes.  First we compute the histograms of object and scene responses, which serve as the evidence to support the need of mining a subset. Then, we propose an effective method to determine this subset of object and scene classes.

\textbf{Distributions of object and scene classes.} With the conditional distributions $p(o|e)$ and $p(s|e)$, and the marginal distribution $p(e)$, we can estimate the marginal distributions of objects and scenes as follows:
\begin{equation}
  p(o)  = \sum_{e} p(o|e)p(e) , \ \ \  p(s)  = \sum_{e} p(s|e)p(e).
\end{equation}
With the above equations, we can estimate the distributions of common objects and scenes in the ChaLearn Cultural Event Recognition dataset. As shown in Figure \ref{fig:os_distribution}, we only choose the top-100 object and scene classes for visual clarity. The distributions of object and scene responses both exhibit ``long tails'', where most of the probability resides in a small subset of these object and scene classes. Many classes only weakly correlate with the event classes. Hence, we select a subset of classes, that are more efficient for storage and subsequent processing.

\textbf{Selection.} The selection of the subset of classes is guided by the following two principles:
\begin{itemize}
  \item Each selected object or scene should only occur in a small number of event classes. We call this property the \emph{discriminative capacity} of a single object or scene.
  \item Each object or scene in the subset should have a low correlation with the others. We call this property the \emph{diversity capacity} of the subset.
\end{itemize}

Based on these principles, we formulate the subset selection as an inference problem on a fully-connected graph, where each node corresponds to an object or scene class, and each edge encodes the correlation between the connected pair of nodes. Each node is associated with a binary hidden variable $h_i \in \{0,1\}$, representing whether the object or scene class is selected or not. Therefore, given a set of object or scene classes $\mathbf{h} = \{h_i\}_{i=1}^N$, we want to minimize the following energy function:
\begin{equation}
\begin{split}
E(\mathbf{h}) & = \sum_{i=1}^N \phi(h_i) + \lambda \sum_{i=1}^N \sum_{j=1,j \neq i}^N \psi(h_i, h_j), \\
& s.t. \sum_{i=1}^N h_i = K
\end{split}
\label{equ:opt}
\end{equation}
where $\phi(h_i)$ is a unary term to represent the cost of selecting the $i^{th}$ class, $\psi(h_i, h_j)$ is a pairwise term to denote the cost of having both $i^{th}$ and $j^{th}$ classes, $K$ is the number of categories to be selected, and $\lambda$ is a weight parameter to balance these two terms (set as 0.5). The unary term is a penalty function to ensure the discriminative capacity of each selected class, and the pairwise term is a penalty function to encourage the diversity of the selected subset.

To meet the requirement of discriminative capacity, the responses of selected objects or scenes should peak for a small subset of events.  Entropy is a natural measure to quantify the peaked nature of a probability distribution. Therefore, we adopt the conditional entropy $H(E|o)$ to represent the discriminative capacity of the $o^{th}$ object, which is defined as follows:
\begin{equation}
H(E|o) = - \sum_e p(e|o) \log_2 p(e|o),
\label{equ:entropy}
\end{equation} 
where $p(e|o)$ is the conditional event distribution given a specific object class, which can be computed from Equation (\ref{equ:likelihood}) and (\ref{equ:prior}) using Bayes' formula, and $M$ is the number of event classes. So, if $h_i = 1$, then $\phi(h_i) = H(E|i)$.

As to a subset's diversity capacity, we need to consider the correlation between pairs of classes. Instead of using low-level features to calculate their similarity, we utilize the conditional probability $P(e|o)$ to measure their correlation. If two object classes would predict similar events, they should have similar conditional probabilities $p(e|o)$. Specifically, if $h_i=1$ and $h_j=1$, then $\psi(h_i,h_j) = < p(e|i), p(e|j)>$.

\begin{algorithm}[t]
\small
  \SetAlgoLined
  \KwData{conditional distribution $p(e|o)$, number: $K$.}
  \KwResult{selected object classes: $\mathcal{O} = \{o_i\}_{i=1}^K$.}
  - Compute the cost of discriminative capacity of each object $\phi(o)$ defined in Equation (\ref{equ:entropy}). \\
  - Initialization: $n \leftarrow 0$, $\mathcal{O} \leftarrow \emptyset$. \\
  \While{$n<K$}{
    1. For each remaining object $o$, update the correlation measure:
    $
       \mathrm{S}(\mathcal{O},o) = \frac{1}{|\mathcal{O}|}\sum_{o_i \in \mathcal{O}} <p(e|o_i), p(e|o)>,
    $\\
    2. Choose the object class : $o^* \leftarrow \mathrm{arg}\min_{o} \phi(o) + \lambda S(\mathcal{O},o)$. \\
    3. Update: $n\leftarrow n+1$, $\mathcal{O} \leftarrow \mathcal{O} \cup \{o^*\} $
    }
  - Return object classes: $\mathcal{O}$.
  \caption{Selecting objects.}
  \label{alg:selection}
\end{algorithm}

In general, the problem in Equation (\ref{equ:opt}) is a combinatorial optimization problem, hard to solve in polynomial time. For special cases of energy functions, the global minimum can be obtained via Graph Cuts \citep{KolmogorovZ04}. We design a greedy selection method as shown in Algorithm \ref{alg:selection}. We first pick the object class that has the lowest conditional entropy. Then we update the correlation of the remaining object classes with the selected one with the average similarity. After this, we choose the object class which simultaneously minimizes the cost of discriminative and diversity capacity and go to the next iteration. The whole iteration will repeat until $K$ objects are selected.


\section{Transferring Deep Representations}
\label{sec:transfer}

\begin{figure*}
\includegraphics[width=\textwidth]{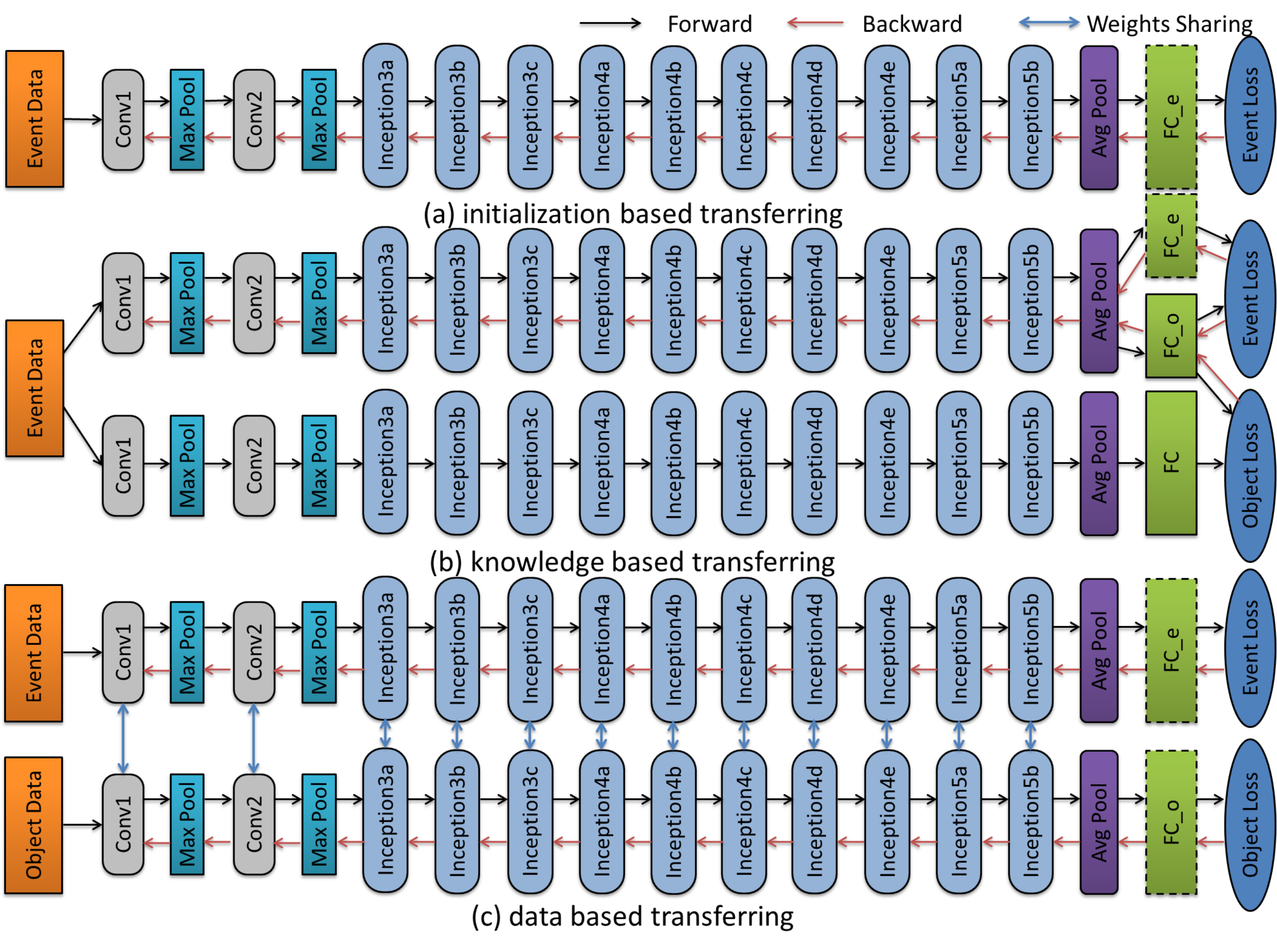}
\caption{Illustration of three transferring techniques that we propose: (a) initialization based transfer, (b) knowledge based transfer, and (c) data based transfer. In initialization based transfer, we use the pre-trained models to initialize the event CNNs and fine-tune them on the target dataset. In knowledge based transfer, we utilize the soft codes produced by object or scene networks to guide the training of event CNNs. In data based transfer, we exploit object or scene training data to jointly learn event and object or scene CNNs in a weight-sharing scheme.}
\label{fig:transfer}
\end{figure*}

After our quantitative investigation of the relation between objects, scenes, and events, and the selection of discriminative object and scene classes, we use these results to transfer the object and scene models (CNNs) to event recognition for the target dataset. The representations of object and scene CNNs are learned to maximize the performance in classifying each image into pre-defined object and scene classes, respectively. The difference of object and scene recognition with event recognition is double: (1) \emph{domain difference}: the distribution of event images is different from those of object and scene images. (2) \emph{target difference}: the final recognition objective is different. Therefore, transferring these deep representations is necessary to deal with the problems of domain and target differences.

Some challenges remain for this transfer. As said, the size of the event recognition datasets is relatively small compared with that of the large-scale object and scene recognition datasets (\emph{e.g.} ImageNet \citep{DengDSLL009} and Places205 \citep{ZhouLXTO14}). Meanwhile, deep CNNs come with millions of parameters and may easily over-fit with limited training samples. Hence, the question of \emph{how to adapt deep representations to new tasks with limited training samples} needs to be explored. As shown in Figure \ref{fig:transfer}, we try to design effective transferring techniques, able to reduce over-fitting to the training data and to improve the generalization capacity of the learned model.

\subsection{Baseline: initialization-based transfer}

Fine-tuning yields a simple yet effective method of adapting deep models trained on a large-scale datasets to a target task with a smaller training dataset. We choose fine-tuning as baseline transferring technique. We call this transferring method as \emph{initialization based transfer}, as we simply copy the weights of object and scene CNN models to the corresponding layers of event models as initialization. Specifically, suppose we have $K$ event classes, we then minimize the cross-entropy loss on the target dataset $D_e$ defined as follows:
\begin{equation}
\ell_C (D_e) = -\sum_{\mathbf{I}_i \in D_e} \sum_{k=1}^K \mathbb{I}(y_i=k) \log p_{i,k},
\label{equ:cls}
\end{equation}
where $(\mathbf{I}_i, y_i)$ is a pair of an image and its label, $D_e$ is the event dataset, $p_{i,k}$ is the $k^{th}$ output of the softmax layer for image $\mathbf{I}_i$. 

In practice, we choose the bn-inception architecture \citep{IoffeS15}, but we make two important modifications when minimizing the above loss to improve the final recognition performance:
\begin{itemize}
  \item We freeze the layers of Batch Normalization in the event CNNs. In the original training of the bn-inception network \citep{IoffeS15}, they estimate the activation mean and variance within each batch and use them to transform these activation values into a standard Gaussian distribution. The data adaptively updating the activation mean and variance contributes to accelerating the CNN's convergence. Yet, it increases the risk of over-fitting when having limited training samples. Therefore, we freeze the mean and variance values as those estimated on the object and scene datasets.
  \item We add a dropout layer before the fully connected layer in the bn-inception network. In the original bn-inception network \citep{IoffeS15}, it turns out Batch Normalization acts as a kind of regularizer, and there is no need for a dropout layer when training on the large-scale ImageNet dataset. However, the size of the event recognition datasets is much smaller than that of ImageNet, and in our experiments, the effect of over-fitting will be more serious without the dropout layer.
\end{itemize} 

This transferring technique simply employs a pre-trained model as initialization and ignores other information during the fine-tuning procedure. Although the fine-tuning process starts with a semantic and stable initialization, it may still suffer from over-fitting, as we shall see in our experiments. Incorporating other relevant tasks into the procedure of fine-tuning will regularize the learning process of event CNNs and thus relieve the over-fitting problem.

\subsection{Knowledge-based transfer}

As shown in Section \ref{sec:corr}, the occurrences of objects and scenes are highly correlated with event classes. The fine-tuning technique simply utilizes object and scene models as initialization but ignores the rich information coming from the present objects and scene background during the fine-tuning. This subsection aims to do that. 

A complicating factor is that we only have event labels on the target dataset, and no object and scene labels. To overcome this, we utilize the pre-trained object and scene CNNs to predict the likelihood of object and scene classes. As shown in Figure \ref{fig:transfer}, we use the soft codes of the object and scene CNNs as supervision to guide the fine-tuning of event CNNs. The advantages of using the soft codes of pre-trained models are two-fold: (1) We do not need to spend much time labeling the images with object and scene annotations. (2) The soft codes also capture the co-occurrence of multiple objects and scenes. Since the knowledge of object and scene CNNs is exploited to obtain the soft codes of images, we call this transferring technique \emph{knowledge-based transfer}. Specifically, during the fine-tuning process, we minimize the following loss function:
\begin{equation}
\ell_{know}(D_e, \mathbf{F}) = \ell_C(D_e) + \alpha \ell_{soft}(D_e, \mathbf{F}),
\label{equ:kd}
\end{equation}
where $\ell_C(D_e)$ is the loss function of event recognition as defined in Equation (\ref{equ:cls}), $\ell_{soft}(D_e, \mathbf{F})$ is the loss of measuring the distance between prediction and the soft codes produced by object and scene CNNs $\mathbf{F}$, and $\alpha$ is a weight to balance these two terms. The loss $\ell_{soft}(D_e, \mathbf{F})$ is formulated as follows:
\begin{equation}
\ell_{soft}(D_e, \mathbf{F}) = -\sum_{\mathbf{I}_i \in D_e} \sum_{k=1}^K q_{i,k} \log f_{i,k},
\end{equation}
where $q_i$ the softmax output of the event network for object or scene prediction with image $\mathbf{I}_i$, and $f_i$ is the softmax output of the pre-trained object or scene models, namely $f_i = \mathbf{F}(\mathbf{I}_i)$.

In this transferring technique, we propose a multi-task learning framework to jointly fine-tune the network weights for event recognition and imitate the object and scene networks to recognize the objects and scene present. This additional imitation task exploits these soft codes to guide the fine-tuning and acts as a kind of regularizer to improve the generalization capacity of event models.

\subsection{Data-based transfer}

In this subsection, we approach learning the event models together with object and scene recognition from a different perspective. In order to incorporate the object and scene recognition into the fine-tuning process, we jointly train CNNs for relevant tasks on different datasets in a weight sharing scheme. We find that this weight sharing scheme is another effective strategy to regularize the fine-tuning of the event network.

Specifically, we simultaneously fine-tune two different networks on two datasets: one network is fine-tuned on the objects and scene subset of ImageNet \citep{DengDSLL009} or Places205 \citep{ZhouKLOT15} for object recognition or scene recognition, and the other one is fine tuned on the event recognition dataset to classify events. As shown in Figure \ref{fig:transfer}, these two networks have their own data layers to handle different datasets, fully-connected layers to deal with different targets, and loss layers to learn the weights for different tasks. The weights of the remaining layers are shared by these two networks. Therefore, during the training process, we minimize the following loss function:
\begin{equation}
\ell_{data}(D_e, D) = \ell_{C}(D_e) + \beta \ell_{C}(D),
\label{equ:data}
\end{equation}
where $\ell_{C}(D_e)$ is the loss function for event recognition on the target dataset as defined in Equation (\ref{equ:cls}), and $\ell_{C}(D)$ is the loss function on the auxiliary dataset for object or scene recognition, $\beta$ is a parameter to keep a balance between these two terms.

This transferring technique aims to utilize useful information hidden in other datasets to help the fine-tuning of event models and we call this method as \emph{data-based transfer}. The weight sharing scheme couples two networks together and jointly updates the network weights during fine tuning. This jointly updating convolutional weights is capable of exploiting the supervision signals from two related datasets to guide network training and can prevent it from over-fitting for any single dataset. The supervision signal from the other dataset acts as a regularizer to improve the quality of fine-tuning on the target dataset.


\section{Event Recognition with OS2E-CNNs}
\label{sec:recognition}

\begin{figure*}
\includegraphics[width=\textwidth]{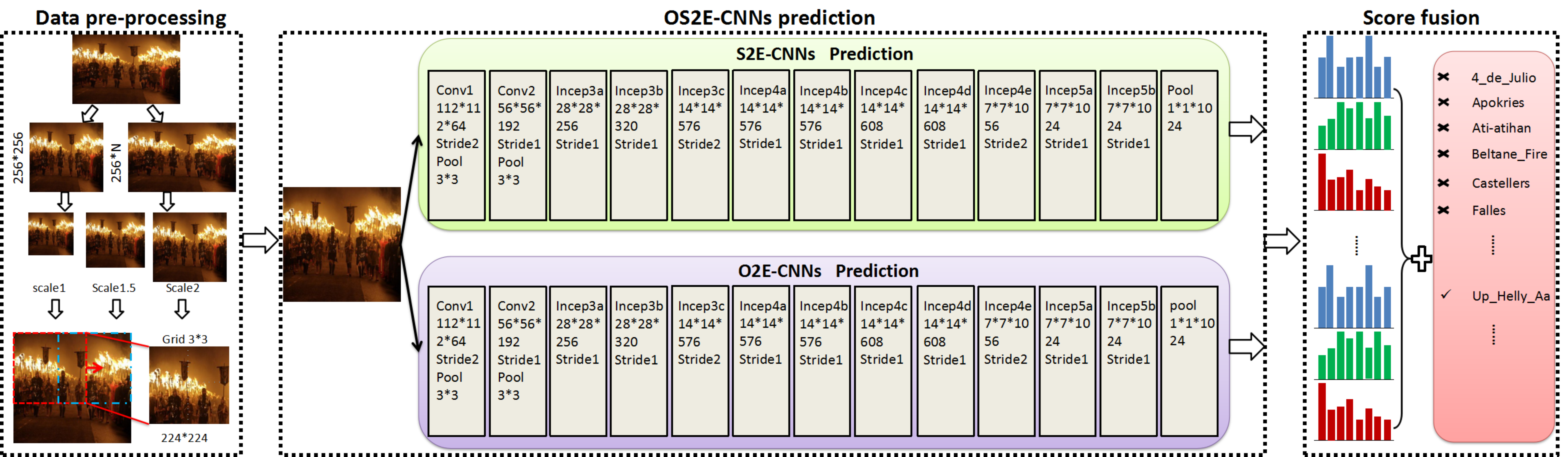}
\caption{\textbf{Pipeline of event recognition with OS2E-CNNs}. We propose a simple yet effective approach for event recognition in a end-to-end manner, which is composed of three steps: (1) data pre-processing, (2) OS2E-CNNs prediction, and (3) score fusion. In the data pre-processing step we transform each image into a set of image regions with a multi-ratio and multi-scale cropping strategy. Then, these crop regions are fed into OS2E-CNNs to predict the score of different event classes. Finally, these scores of different image regions are fused to yield the final recognition result.}
\label{fig:pipeline}
\end{figure*}

After the introduction of transferring techniques from object and scene models to event CNNs, we are ready to describe how to deploy these fine-tuned CNNs for event recognition.

The pre-trained object and scene CNNs are fine-tuned for event recognition on the event dataset and we call these learned networks {\bf OS2E-CNNs}, which is short for \emph{convolutional neural networks transferred from objects and scenes to events}.  As shown in Figure \ref{fig:pipeline}, the OS2E-CNNs are composed of two streams: (1) O2E-CNNs: event CNNs transferred from object models and object datasets. (2) S2E-CNNs: event CNNs transferred from scene models and scene datasets. In order to efficiently deploy OS2E-CNNs, our current implementation treats OS2E-CNNs as ``end-to-end predictors'' without training explicit classifiers such as SVMs. As illustrated in Figure \ref{fig:pipeline}, our event recognition pipeline involves three steps: (1) data pre-processing, (2) network prediction, and (3) score fusion.

\textbf{Data pre-processing.}  We conduct data augmentation techniques to enrich the image samples. First, we propose to resize the image with two different aspect ratios: (1) keeping the image aspect ratio fixed and setting the smaller side to $256$, (2) resizing the image to $256 \times 256$. Second, to deal with scale variations, we change the image resolutions with the smaller side as $384$, $512$ or the whole size as $384 \times 384$, $512 \times 512$. Finally, we densely crop $224 \times 224$ regions in a grid of $3 \times 3$ from these images. Hence, we crop a total number of $2 \times 3 \times 9=54$ regions from a single image.

\textbf{Network prediction.}  We first subtract their mean pixel value from the cropped image regions and then feed the results into the O2E-CNN and S2E-CNN independently. The O2E-CNN and S2E-CNN produce prediction scores at the softmax layer. These score vectors represent the likelihood of event categories for each image region.

\textbf{Score fusion.}  We combine the scores of the different networks for different crops. First, for each cropped region $\mathbf{R}$, the prediction score of the OS2E-CNN is a weighted average of O2E-CNN and S2E-CNN results:
\begin{equation}
S_{os}(\mathbf{R}) = \alpha_o S_o(\mathbf{R}) + \alpha_s S_s(\mathbf{R}),
\end{equation}
where $S_o(\mathbf{R})$ and $S_s(\mathbf{R})$ are the score vectors of the O2E-CNN and S2E-CNN for region $\mathbf{R}$.  $\alpha_o$ and $\alpha_s$ are their fusion weights and in the current implementation are equal. Then, for the whole image $\mathbf{I}$, the prediction score is obtained by fusing across these cropped regions:
\begin{equation}
S_{os}(\mathbf{I})  = \sum_{\mathbf{R}_i \in \mathbf{I}} S_{os}(\mathbf{R}_i).
\end{equation}


\section{Experiments}
\label{sec:exp}

In this section, we describe the detailed experimental setting and report the performance of our method. In particular, we first introduce the datasets used for evaluation and their corresponding experimental setup. Next, we describe the implementation details of how to transfer object and scene models to event recognition. Then we explore the multi-ratio and multi-scale cropping strategy. After this, we study different aspects of our proposed object and scene selection method. Meanwhile, we compare and analyze the performance of the three transferring techniques. We also compare the performance of our method with that of winners of the ICCV15 ChaLearn Looking at People (LAP) challenge. Furthermore, we fix the parameter settings and perform experiments on two other event recognition datasets. Finally, we give examples for which our method fails to predict the correct labels.

\subsection{Datasets and evaluation protocol}

In our experiment, we choose three challenging event recognition datasets:  (1) the ChaLearn Cultural Event Recognition dataset \citep{ICCV_LAP}, (2) the Web Image Dataset for Event Recognition (WIDER) \citep{XiongZLT15}, and (3) the UIUC Sports Event dataset \citep{LiF07}.

The ChaLearn Looking at People (LAP) challenge~\footnote{\url{http://gesture.chalearn.org/}} is becoming an important contest focusing on human pose estimation, action and gesture recognition, human face analysis, and cultural event recognition \citep{ECCV_LAP,CVPR_LAP,ICCV_LAP}. The ICCV15 ChaLearn LAP challenge provides a large dataset for cultural event recognition. This dataset contains images collected with two image search engines (Google Images and Bing Images). There are 100 event classes in total (99 event classes and 1 background class), from different countries. The whole dataset is divided into three parts: development data (14,332 images), validation data (5,704 images), and evaluation data (8,669 images). The principal quantitative measure is based on the precision-recall curve. We average these per-class average precision (AP) values across all event classes and employ the mean average precision (mAP) as the final ranking criteria.

We perform two experiments with different settings on the ChaLearn Cultural Event Recognition dataset. The first is the \textbf{validation setting}, where we train our CNN models on the development data and test our CNN models on the validation data. As we can not access the labels of the test data, we explore different configurations of our method to determine optimal parameters. The second experiment uses a \textbf{challenge setting}, where we merge the development data (14,332 images) and validation data (5,704 images) into a single training dataset (20,036 images) and re-train our OS2E-CNNs with this new training dataset as for the validation setting. We sent our recognition results to the challenge organizers and obtained the final performance back. 

The recently introduced Web Image Dataset for Event Recognition (WIDER)~\footnote{\url{http://personal.ie.cuhk.edu.hk/~xy012/event_recog/WIDER/}} is probably the largest event recognition benchmark for still images. In the current version, there are 50,574 images in total, annotated with event labels from 61 categories. The whole dataset is divided into 25,275 training images and 25,299 testing images. The evaluation measure is based on the mean recognition accuracy across all the event classes. Different from the ChaLearn Cultural Event Recognition dataset, WIDER focuses on more everyday types of events, such as parade, dancing, meeting, and press conference. Complementary as it is to the ChaLearn dataset, evaluation on WIDER allows us to further verify the effectiveness of our method.

The UIUC Sports Event dataset \footnote{\url{ http://vision.stanford.edu/lijiali/event_dataset/}} probably was the first event recognition benchmark for still images. It is composed of 8 sports categories from the Internet: bocce, badminton, croquet, rock climbing, snowboarding, sailing, and polo. The number of images in each event category ranges from 137 to 250. Following the original evaluation setting, we randomly select 70 images for each event class as training samples and 60 images as testing samples. The final evaluation is based on the mean recognition accuracy across the 8 event classes. Compared with the two previous datasets, this event dataset focuses on the sports and has a smaller size. Performance levels have already quite saturated (around 95\%) and it will be difficult to achieve improvements over the state-of-the-art. Nevertheless, this dataset can help to verify the effectiveness of our proposed transferring techniques if our method is still able to boost the final recognition performance.

\subsection{Implementation details}

In this subsection we introduce the implementation details of our transferring methods. Data augmentation is a technique to produce more training samples by perturbing an image with transformations but leaving the underlying class unchanged. This technique is quite effective to reduce the effect of over-fitting and improve the generalization ability of CNN models, in particular for datasets with limited training samples. Specifically, during the fine tuning of OS2E-CNNs, we first resize each training image to $256 \times 256$. At each iteration, we randomly crop a region from the whole image. To deal with scale and aspect ratio variations, we design a multi-scale cropping strategy, where the cropped width $w$ and height $h$ are randomly picked from $\{256, 224, 192, 160, 128\}$. Then these cropped regions are resized to $224 \times 224$ for network training.  These cropped regions also undergo random horizontal flipping. In general, the network weights are learned using the mini-batch stochastic gradient descent with momentum (set to 0.9). At each iteration, a mini-batch of 256 images is constructed by random sampling. The dropout ratio for the added dropout layer is set as $0.7$ to reduce over-fitting.  As we pre-train the network weights with the ImageNet and Places205 models, we set a smaller learning rate for fine-tuning the OS2E-CNN, and we initialize it as $0.01$. After this, we decrease the learning rate every $K$ iterations and the whole training procedure stops at $2.5K$ iterations. $K$ is related to the training set size and we set it to $5,000$ for the validation setting of ChaLearn Cultural Event Recognition, $7,000$ for the challenge setting of this dataset, $10,000$ for WIDER, and $300$ for the UIUC Sports Event dataset.

\subsection{Exploration of testing strategy}

\begin{table}
\caption{Performance of different cropping strategies on the ChaLearn Cultural Event Recognition dataset under the {\bf validation setting}. We use two image aspect ratios and three different scales. These cropped regions from different resolutions and scales are complementary to each other.}
\label{tbl:cropping}
\resizebox{\linewidth}{!}{
\begin{tabular}{|c|c|c|c|c|}
\hline
Ratio & Scale & O2E-CNNs & S2E-CNNs & OS2-E-CNNs \\ \hline \hline
\multirow{4}{*}{\rotatebox{90}{$256 \times N$}} & Scale 1 & 82.1\% &  80.5\% & 84.2\% \\
\cline{2-5} 
& Scale 1.5 & 81.8\% & 81.1\% & 84.1\% \\
\cline{2-5}
& Scale 2 & 77.2\% & 76.1\% & 79.4\% \\
\cline{2-5}
& combine & 83.4\% & 82.8\% & \textbf{85.3\%} \\ \hline \hline
\multirow{4}{*}{\rotatebox{90}{$256 \times 256$}} & Scale 1 & 80.4\% & 78.2\% & 82.8\% \\
\cline{2-5} 
& Scale 1.5 & 82.4\% & 80.8\% & 84.3\% \\
\cline{2-5}
& Scale 2 & 81.7\% & 80.5\% & 83.5\% \\
\cline{2-5}
& combine & 83.2\% & 82.0\% & \textbf{85.0\%} \\ \hline \hline
\multirow{4}{*}[-1mm]{\rotatebox{90}{Combine}} & Scale 1 & 82.0\% & 80.3\% & 84.1\% \\
\cline{2-5} 
& Scale 1.5 & 83.2\% & 82.1\% & 85.0\% \\
\cline{2-5}
& Scale 2 & 81.2\% & 80.3\% & 83.0\% \\
\cline{2-5}
& combine & 83.9\% & 83.0\% & \textbf{85.6\%} \\ \hline
 \end{tabular}
 }
\end{table}

We begin our experiment by exploring the effectiveness of the multi-ratio and multi-scale cropping strategy proposed in Section \ref{sec:recognition}. Specifically, in this experiment, we use the ChaLearn Cultural Event Recognition dataset under the validation setting and choose initialization-based transfer to learn the event CNN model. The results are reported in Table \ref{tbl:cropping} and we notice that the strategy of multi-ratio and multi-scale cropping is helpful for improving recognition performance. 

First, given a fixed image aspect ratio, we resize the image to three different scales: (1) the original scale, (2) 1.5 times that scale, and (3) double the scale, and at each scale, we crop $3 \times 3$ image regions of size $224 \times 224$. The performance of using three different scales is improved to 85.3\% for the OS2E-CNN compared with the original performance 83.4\%. This improvement suggests the multi-scale cropping method can handle scale variations of the test images. Second, we choose two aspect ratios for the testing images ($256 \times N$ vs. $256 \times 256$) and they obtain similar performance (85.3\% vs. 85.0\%). We fuse the recognition results of these two aspect ratios and can boost the recognition performance to 85.6\%. This improvement may be ascribed to an aspect ratio difference among test images. This aspect ratio jittering technique may be helpful to handle this issue. In summary, the strategy of multi-ratio and multi-scale cropping is simple yet quite effective for improving the performance of event recognition from still images. In the remainder of this section, we will use this cropping technique for other experimental explorations. 

\begin{figure*}
\centering
\includegraphics[width=0.245\textwidth]{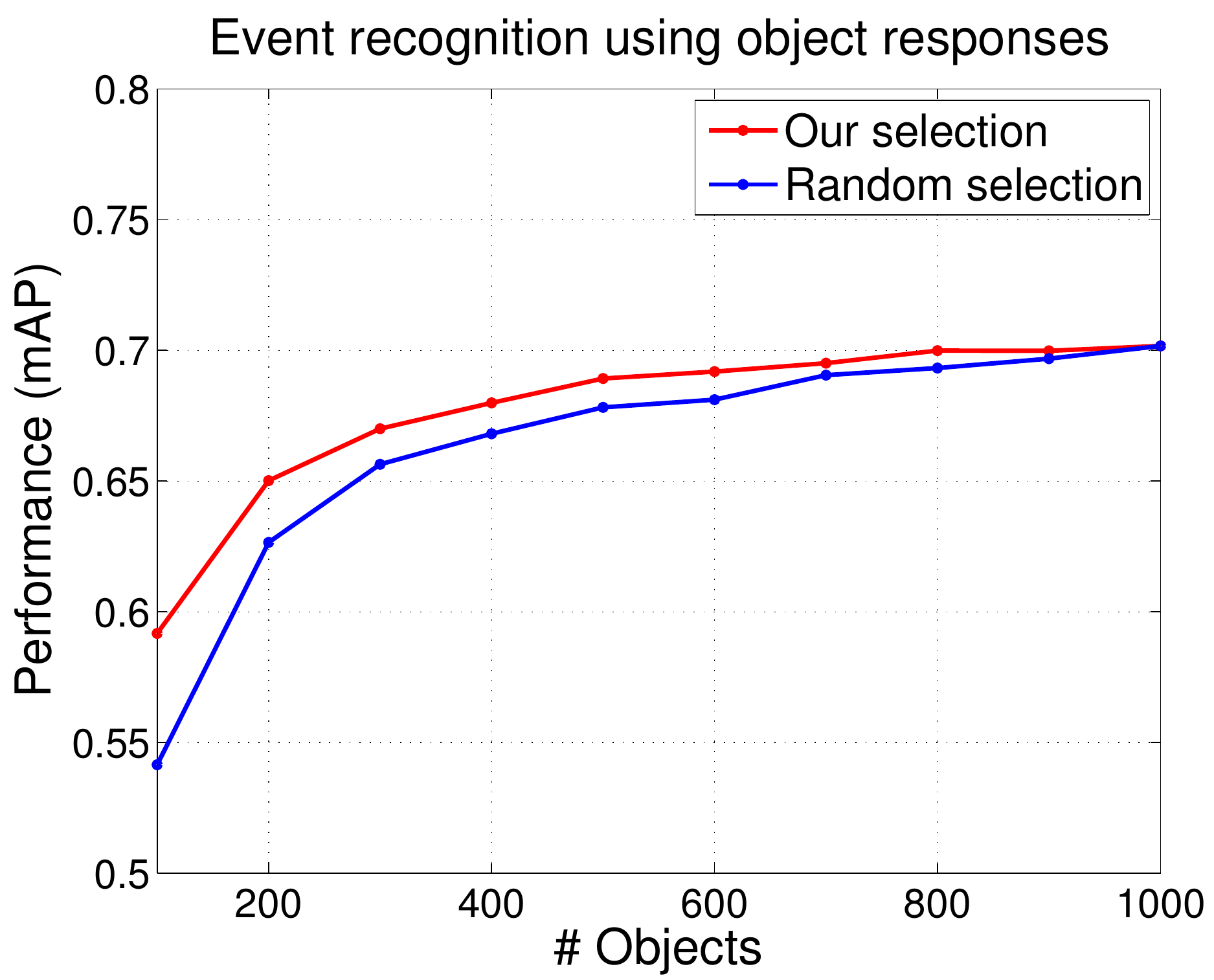}
\includegraphics[width=0.23\textwidth]{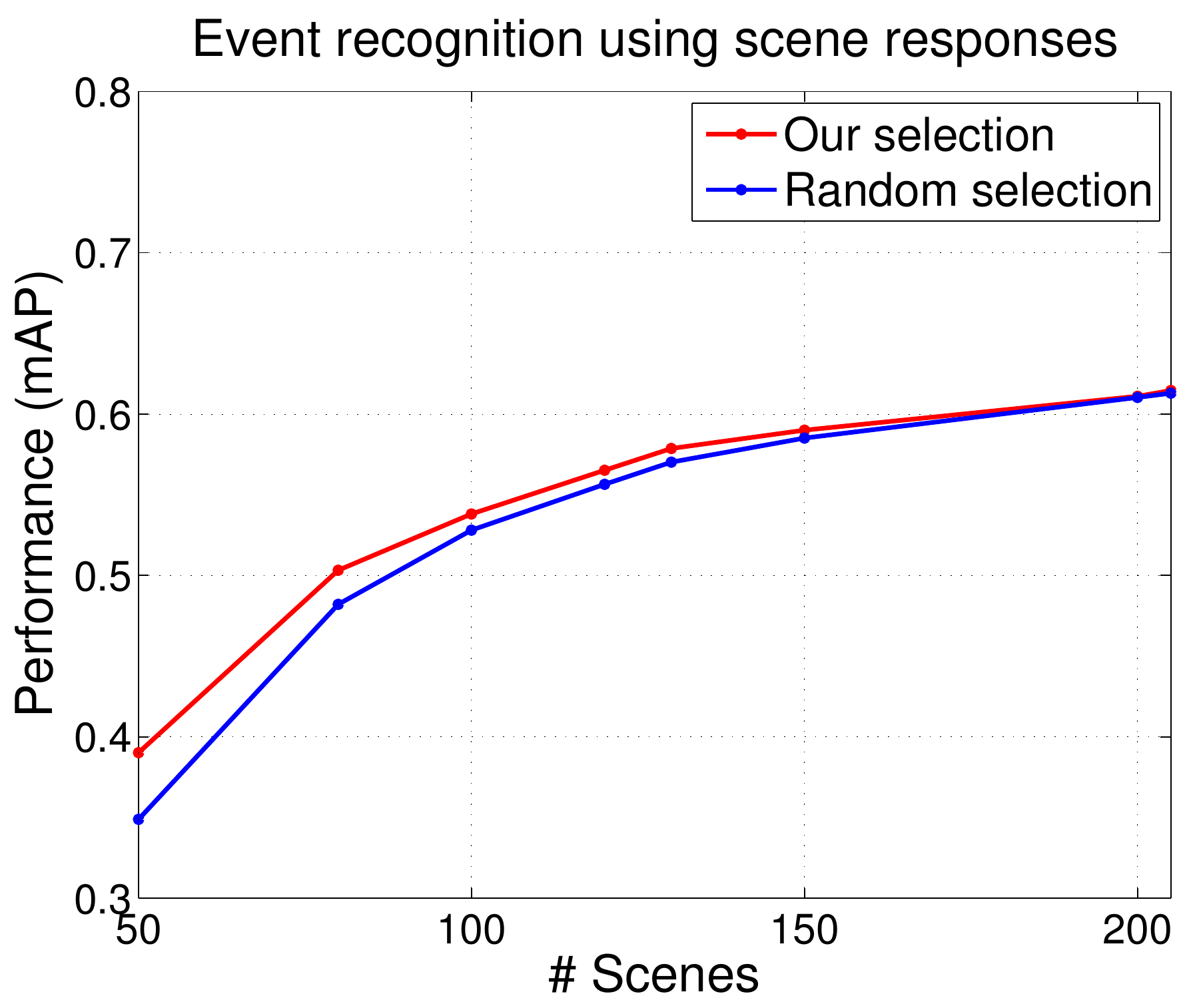}
\includegraphics[width=0.25\textwidth]{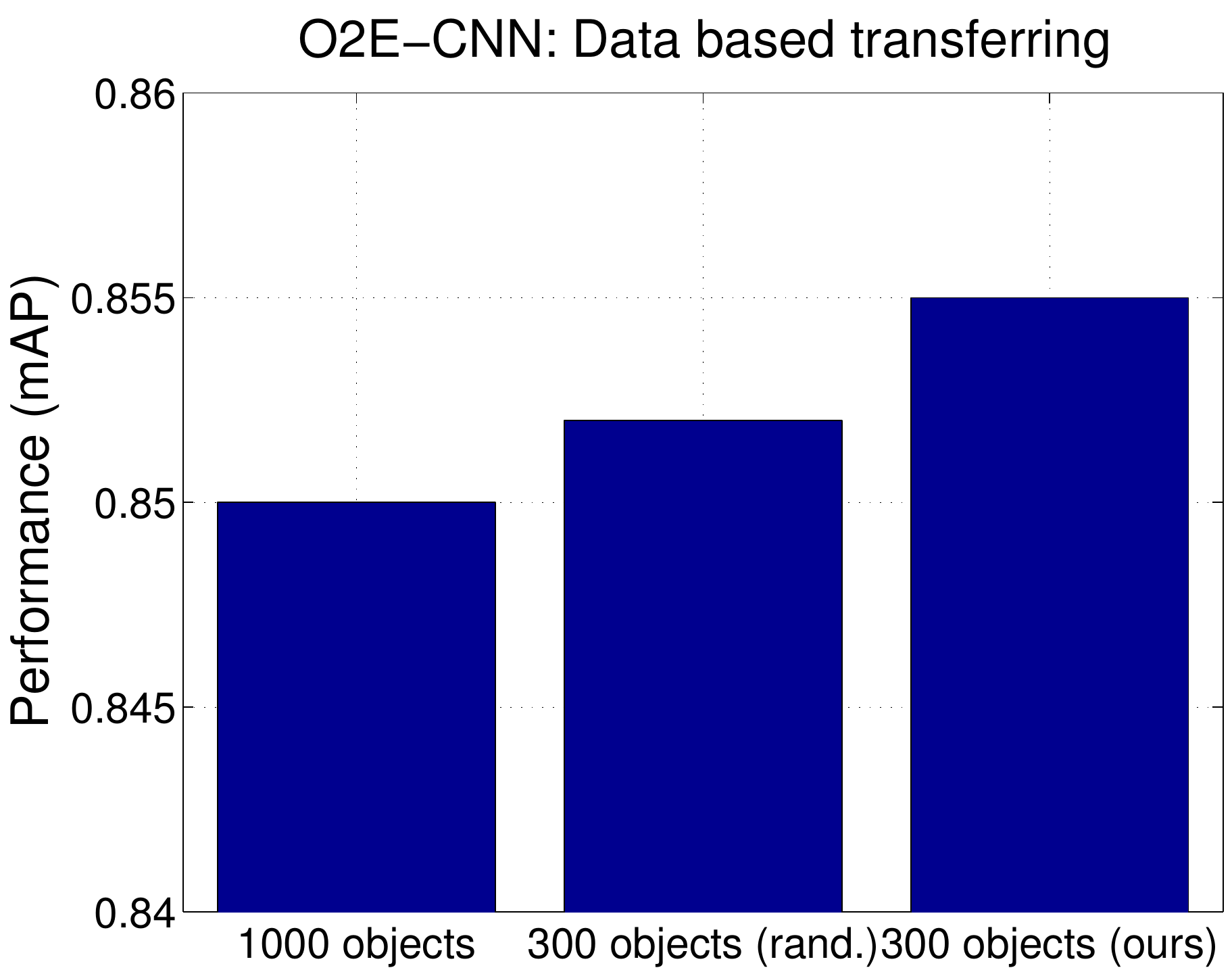}
\includegraphics[width=0.25\textwidth]{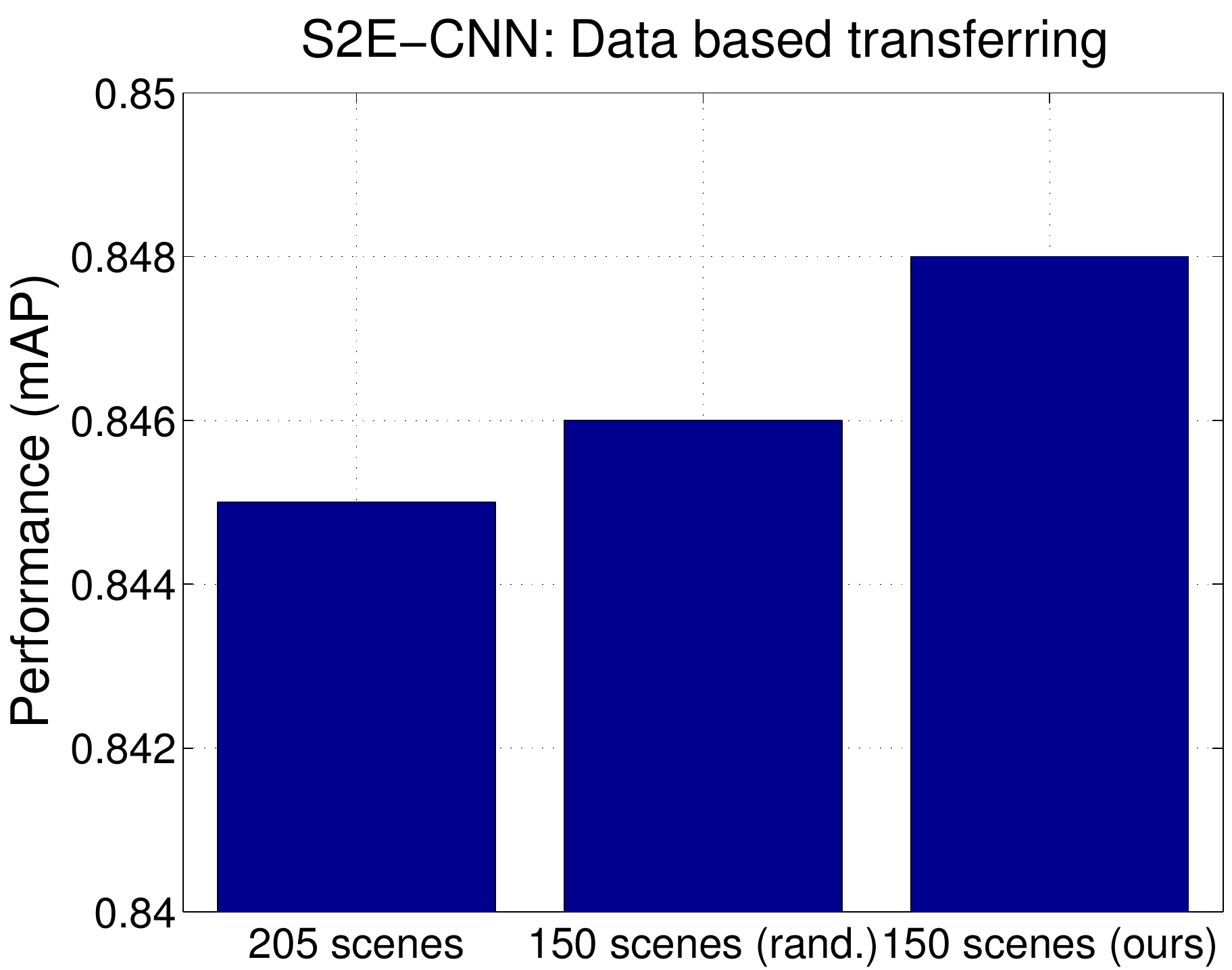}
\caption{Exploration of the object and scene selection algorithm on the ChaLearn Cultural Event Recognition dataset under the {\bf validation setting}. We first use the object and scene responses as features and resort to a linear SVM for classification. The results are shown in the left and we compare with the random selection algorithm. Then, we choose the data-based transfer technique to study the effect of selecting a subset of objects and scenes. The results are shown in the right and we compare with the random selection and usage of all classes.}
\label{fig:selection1}
\end{figure*}

\begin{figure*}
\includegraphics[width=\textwidth]{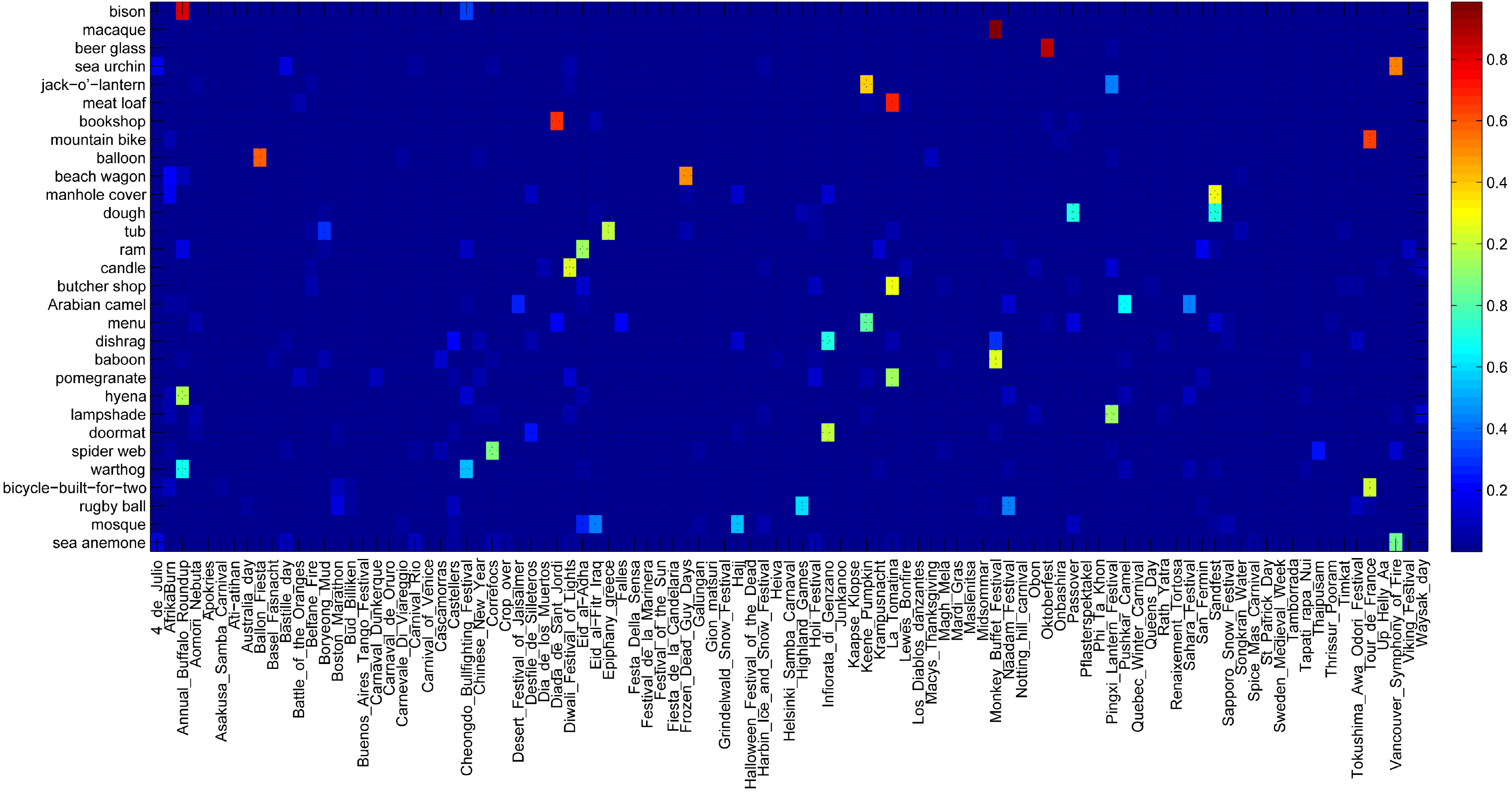}
\includegraphics[width=\textwidth]{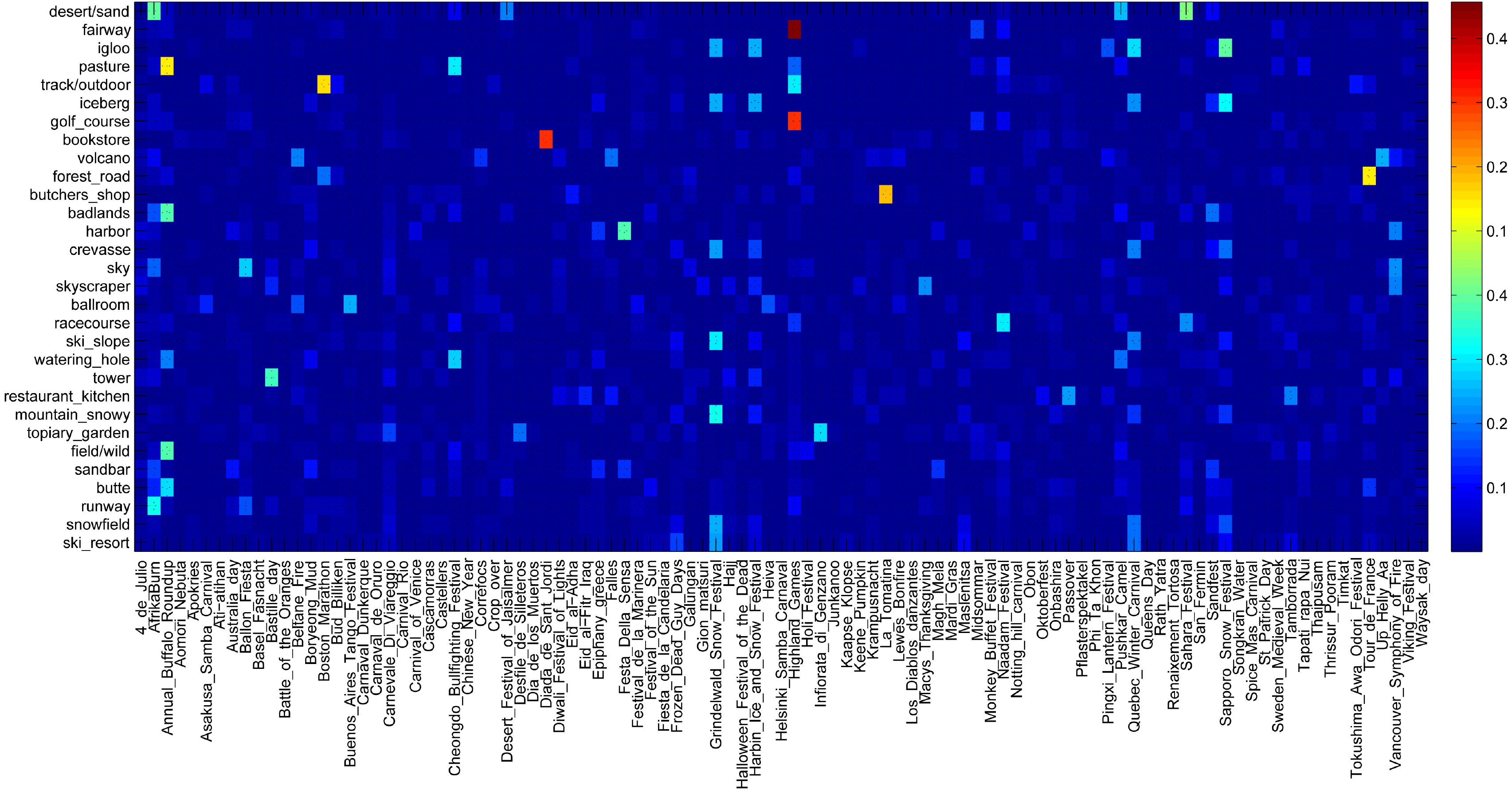}
\caption{Visualization of the conditional probabilities $p(e|o)$ and $p(e|s)$ of selected objects and scenes for the ChaLearn Cultural Recognition dataset. For visual clarification, we plot the top 30 objects (top row) and scenes (bottom row) in the order they are selected. From these results, we see that our selection method is able to find a subset of discriminative and diverse objects and scenes.}
\label{fig:selection2}
\end{figure*}

\subsection{Evaluation of object and scene selection}

\begin{figure*}
\centering
\includegraphics[width=0.24\textwidth]{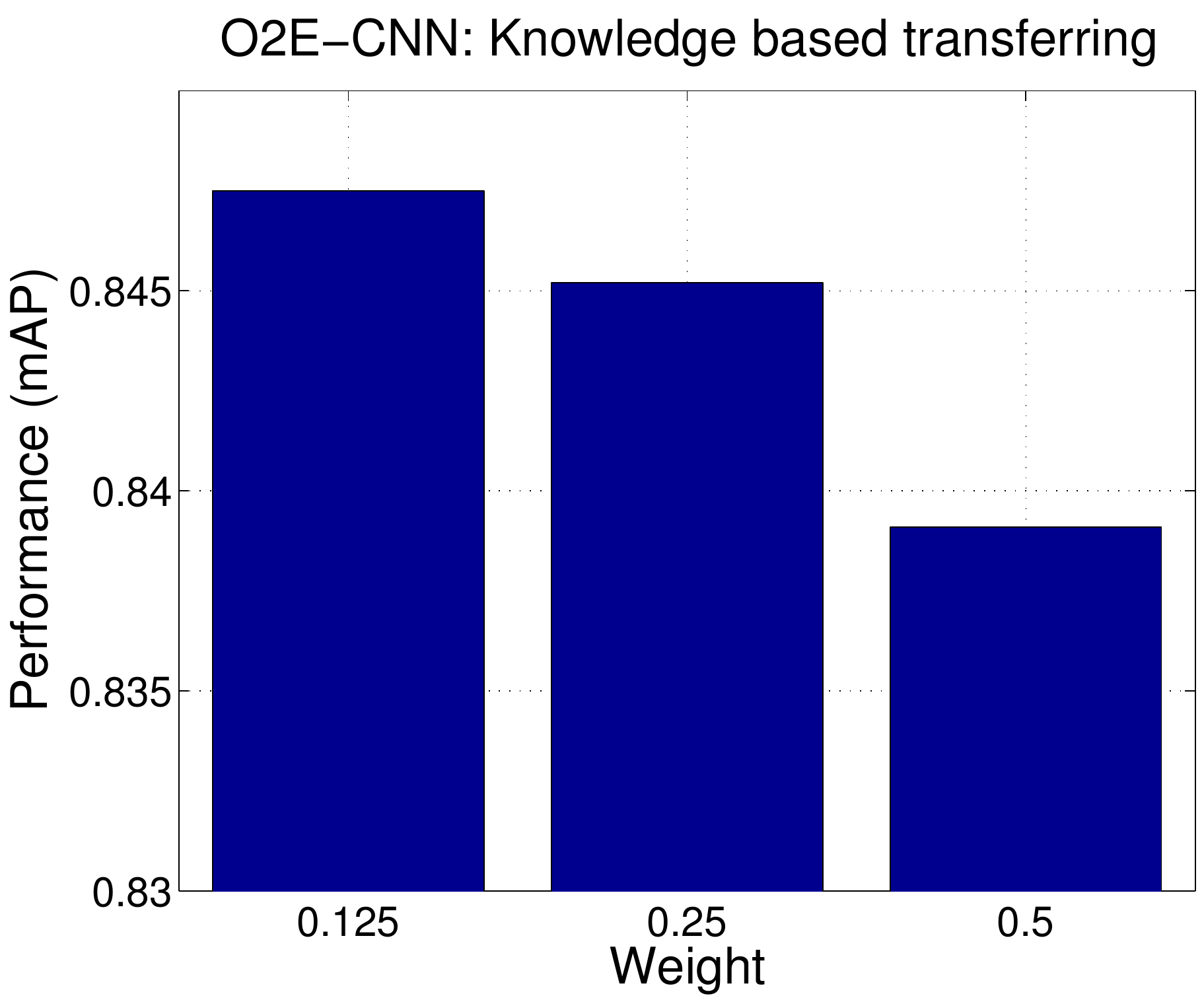}
\includegraphics[width=0.24\textwidth]{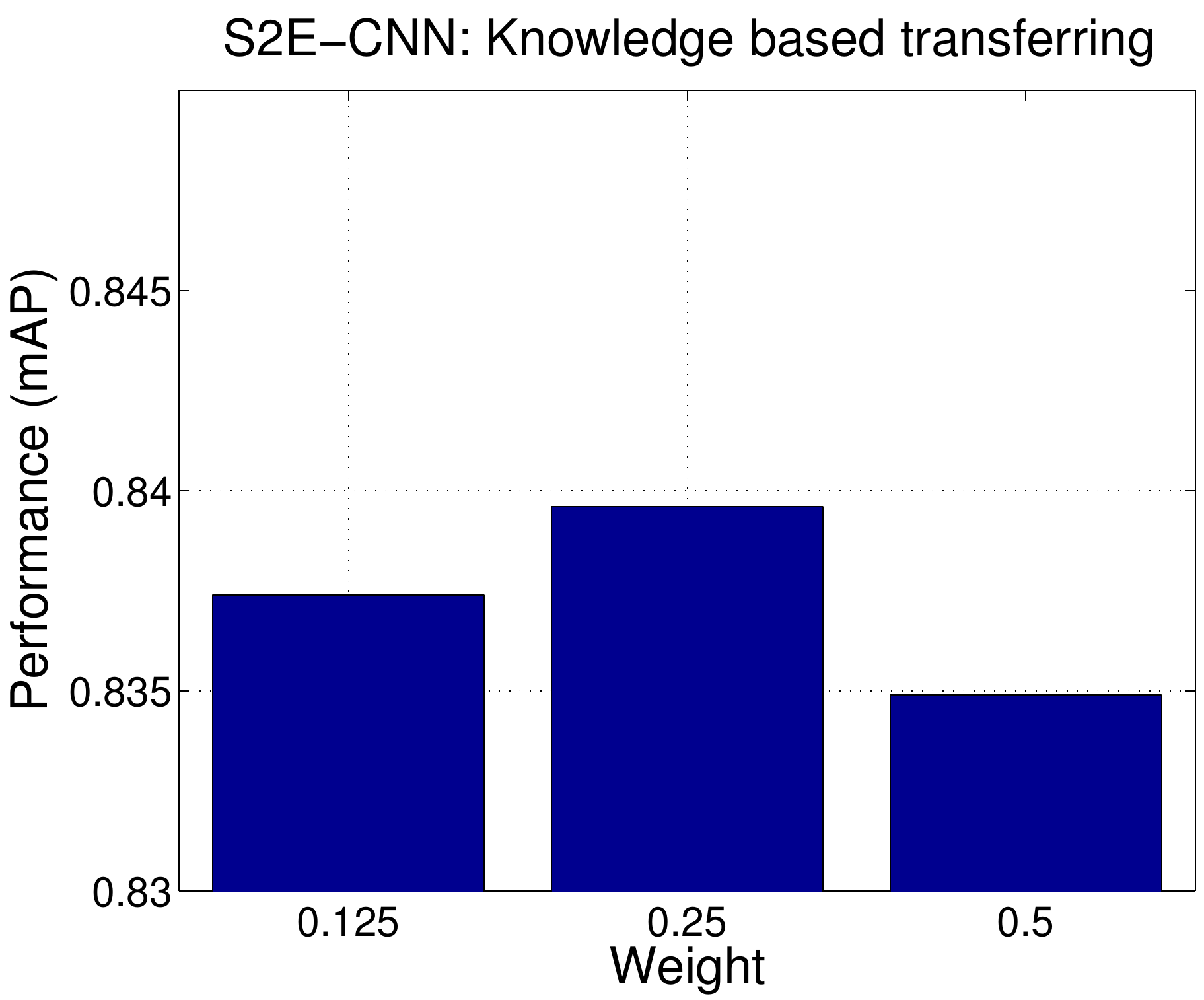}
\includegraphics[width=0.24\textwidth]{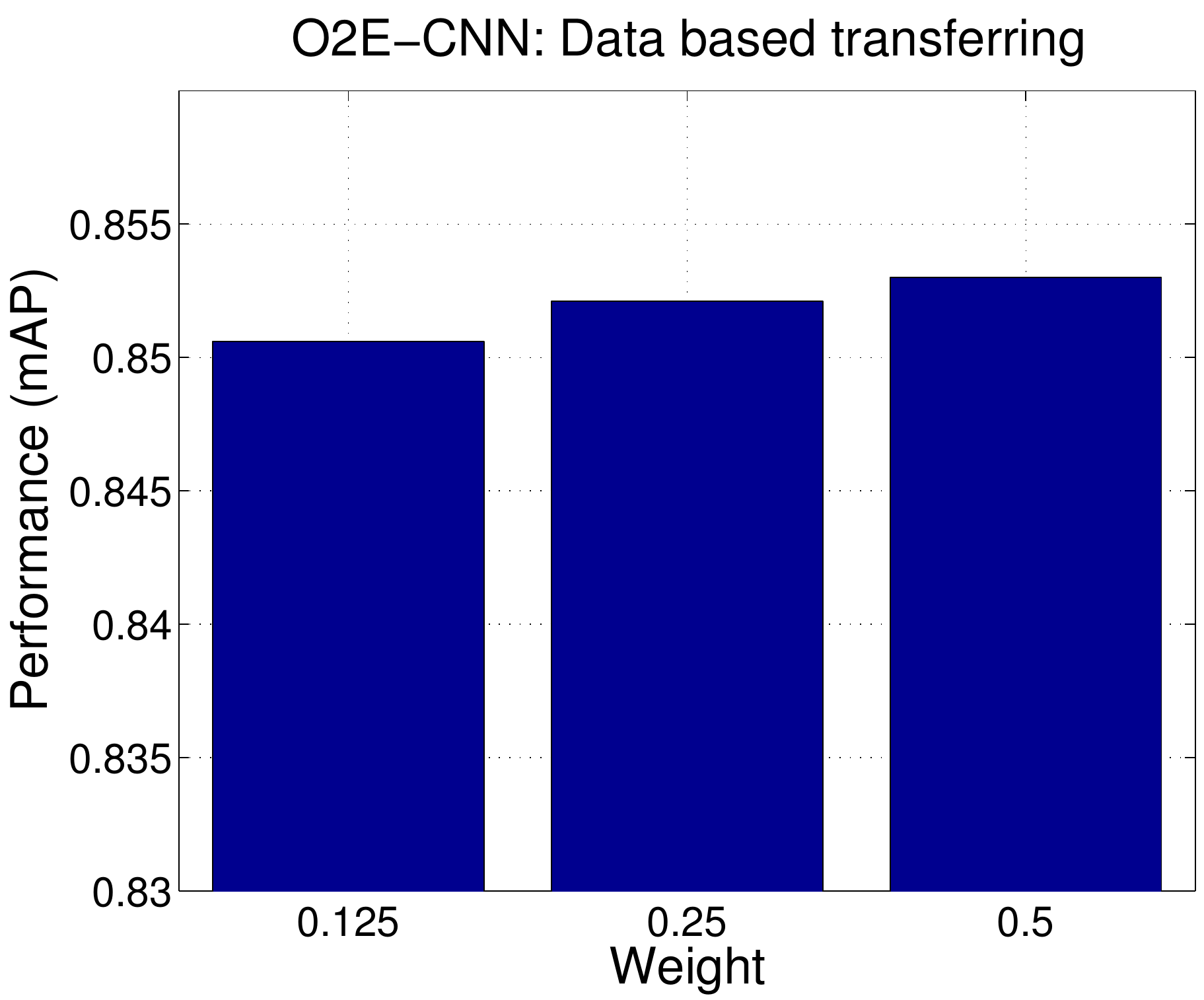}
\includegraphics[width=0.24\textwidth]{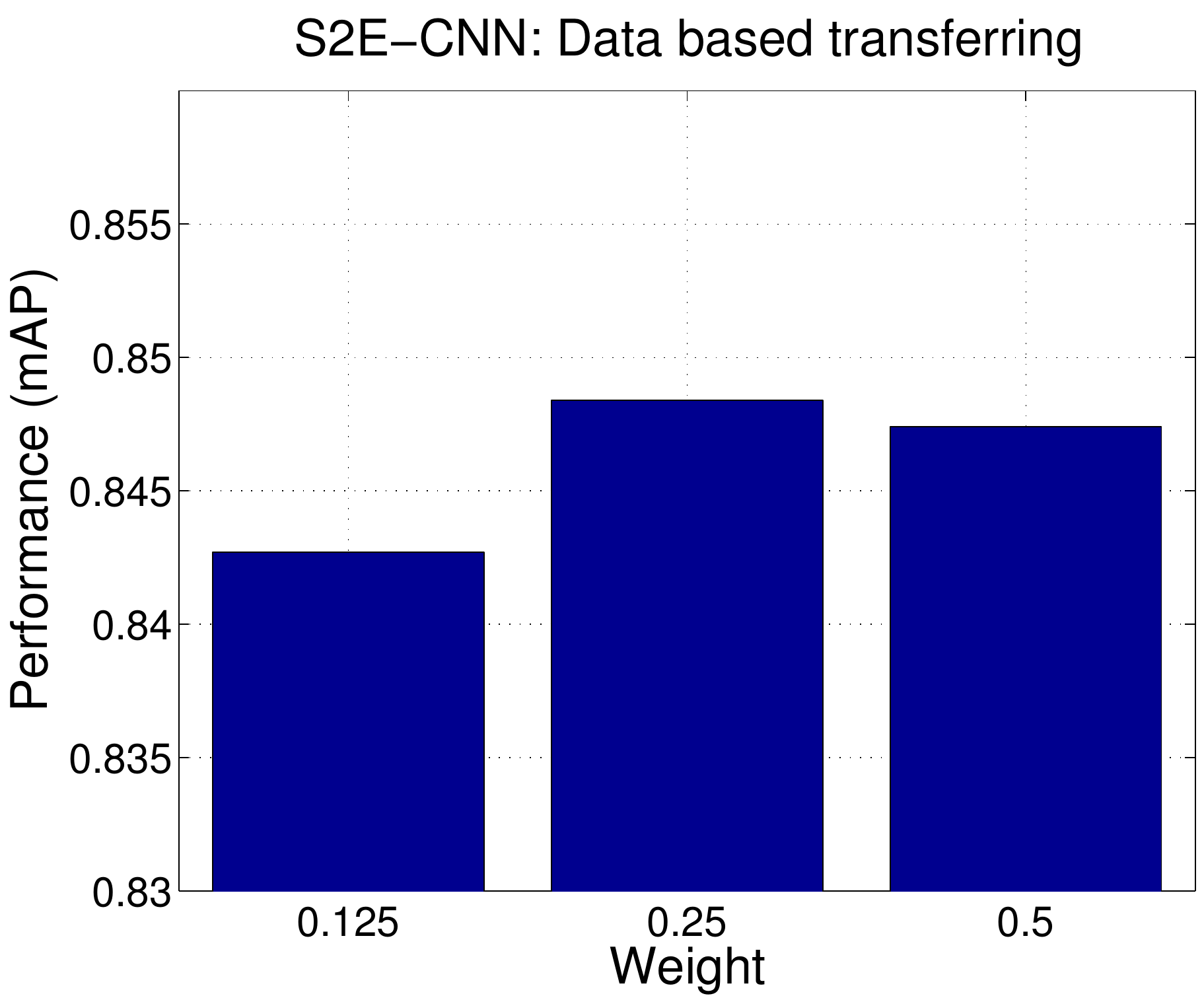}
\caption{Performance of different weights of auxiliary tasks on the ChaLearn Cultural Event Recognition dataset under the {\bf validation setting}. We study both knowledge based transferring and data based transferring methods and aim to find the optimal parameter setting.}
\label{fig:weight}
\end{figure*}

In this subsection, we aim to verify the effectiveness of the object and scene selection method proposed in Section \ref{sec:selection}, on the ChaLearn Cultural Event Recognition dataset under the validation setting. The experimental results are reported in Figure \ref{fig:selection1} and Figure \ref{fig:selection2}. 

First, we treat the object and scene responses $\Phi^o(\mathbf{I})$ and $\Phi^s(\mathbf{I})$ as features and train a linear SVM to classify event classes. The experimental results are reported in the left of Figure \ref{fig:selection1}, where we plot the recognition performance with different numbers of selected object and scene classes. We compare our method with a baseline of random selection and our selection algorithm outperforms that baseline. The performance gap is larger when the selected number is small and this gap becomes smaller when more object and scene classes are selected. As more classes are selected, those discriminative object and scene classes are more easily picked by a random sampling method. Our method is quite effective and can select a relatively small subset of classes, that achieves the $95\%$ performance of using all classes. For instance, using 300 object classes obtains a performance of around $67\%$ and 150 scene categories get a recognition result of around $59\%$. Hence, for fast processing, we fix the selection number as $300$ for objects and $150$ for scenes in the remaining experimental explorations.

Then, we study the effectiveness of selecting a subset of object and scene classes with data-based transfer. We compare with two other methods: (1) using all object and scene classes, (2) selecting 300 object classes and 150 scene classes with random selection. The results are summarized in the right of Figure \ref{fig:selection1}. For the O2E-CNNs, using all the objects (1,000 classes) achieves the performance of $85.0\%$, which is lower than that of employing $300$ object classes, no matter which selection method is adopted. Using smaller number of object classes may contribute to the convergence of training O2E-CNNs. Comparing the performance of random selection and our proposed selection method, this confirms the effectiveness of considering discriminative and diversity capacity during the selection process. For S2E-CNNs, similar results to O2E-CNNs are observed, and our proposed selection algorithm outperforms the other two methods. In summary, our selection algorithm not only requires less additional training images (300 classes vs. 1000 classes), but also helps the fine-tuned model to better generalize on the test dataset. 

Finally, to further explore the details of our selection algorithm, we visualize the conditional probabilities $p(e|o)$ and $p(e|s)$ of selected objects and scenes in Figure \ref{fig:selection2}. Here we only plot the top 30 selected classes, listed in the same order as they were selected. We notice that our selection algorithm is able to choose discriminative object and scene classes, while keeping their diversity. For example, the first selected object class is \texttt{bison}, which is good to discriminate the event \texttt{annual bufallo roundup} from other classes, and the first selected scene class is \texttt{desert/sand}, a strong indicator for event class \texttt{afrika burn} and \texttt{sahara festival}.  Meanwhile, our selected object and scene classes appear across different event classes and ensure the subset diversity.

\subsection{Exploration of auxiliary task weights}

We now explore the performance of different weight settings for multi-task based transfer: (1) knowledge-based transfer, and (2) data-based transfer, on the ChaLearn Cultural Event Recognition dataset under the validation setting. These two transfer approaches aim to exploit a multi-task learning framework to leverage extra tasks as regularizers to help fine-tune networks. An important parameter in this multi-task framework is the weights of auxiliary tasks, namely parameters $\alpha$ and $\beta$ in Equation (\ref{equ:kd}) and Equation (\ref{equ:data}), respectively.

We first study the effect of weight $\alpha$ in knowledge-based transfer. As our goal is to perform event recognition, we constrain the weight of the auxiliary task to be less than $0.5$. Specifically, we choose three different weights $0.125$, $0.25$, and $0.5$, and the experimental results of both the O2E-CNN and S2E-CNN are shown in Figure \ref{fig:weight}. For the O2E-CNN, smaller weights achieve better performance and the weight of $0.125$ gets the best performance of around $84.7\%$. However, for the S2E-CNN, the best weight is $0.25$, where it obtains a performance of around $83.9\%$. The difference between O2E-CNN and S2E-CNN may be ascribed to the fact that the scene network performs poorer than the object one. Thus, the effect of over-fitting is more serious for the S2E-CNN than the O2E-CNN and we need to set a higher weight for the auxiliary task to better regularize the training of event CNNs.

We then compare the performance of O2E-CNNs and S2E-CNNs with different weight settings for data-based transfer. The results are reported in Figure \ref{fig:weight}. From these results, we see that the performance of data-based transfer is less sensitive to the weight setting, where weight $0.125$ achieves the lowest performance, and the weights $0.25$ and $0.5$ obtain a similar performance. Hence, in the remaining experimental exploration, we fix the weight of the auxiliary task to $0.5$ for both the O2E-CNN and S2E-CNN.

\subsection{Comparison of transferring techniques}

\begin{figure*}
\centering
\includegraphics[width=0.32\textwidth]{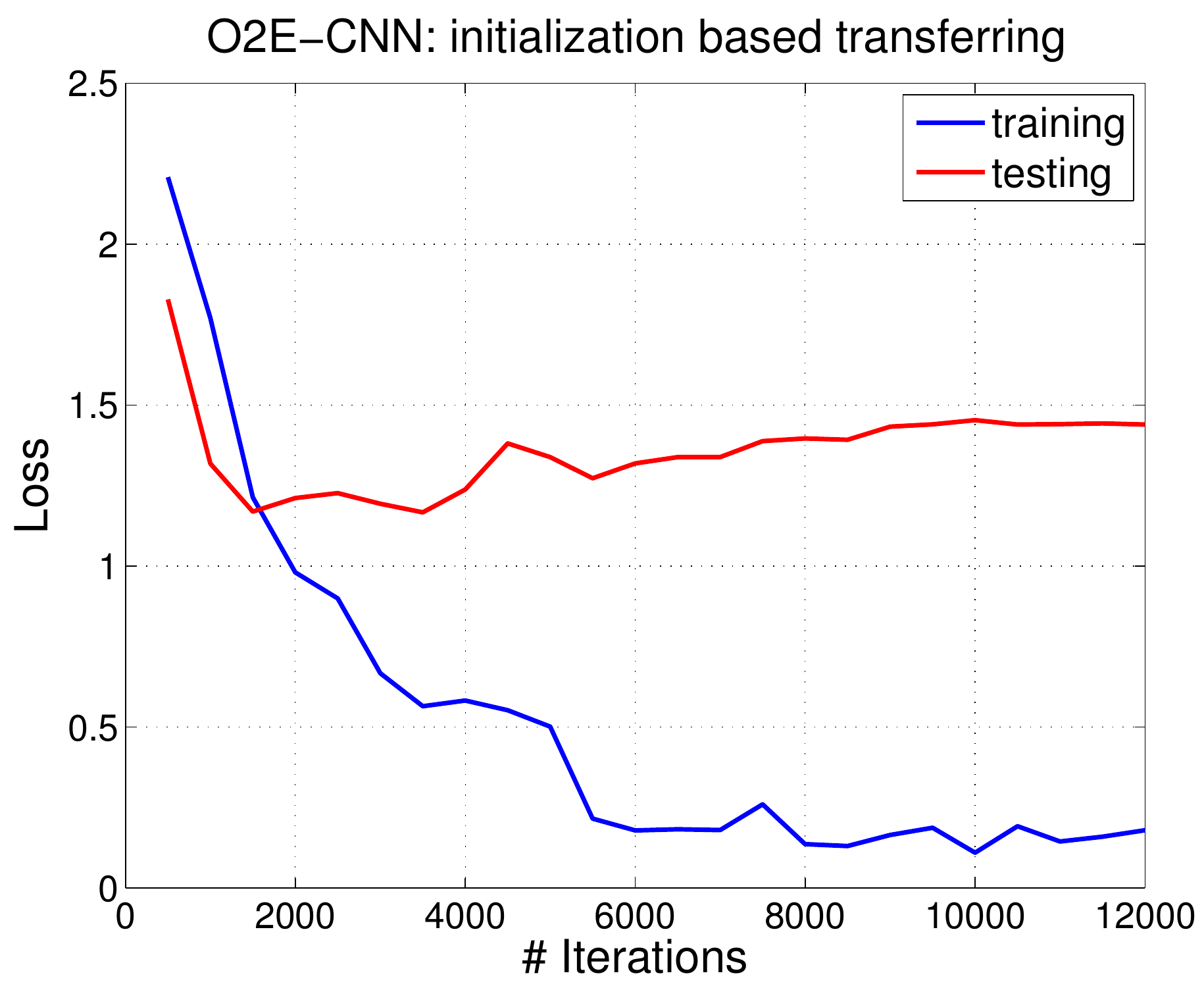}
\includegraphics[width=0.32\textwidth]{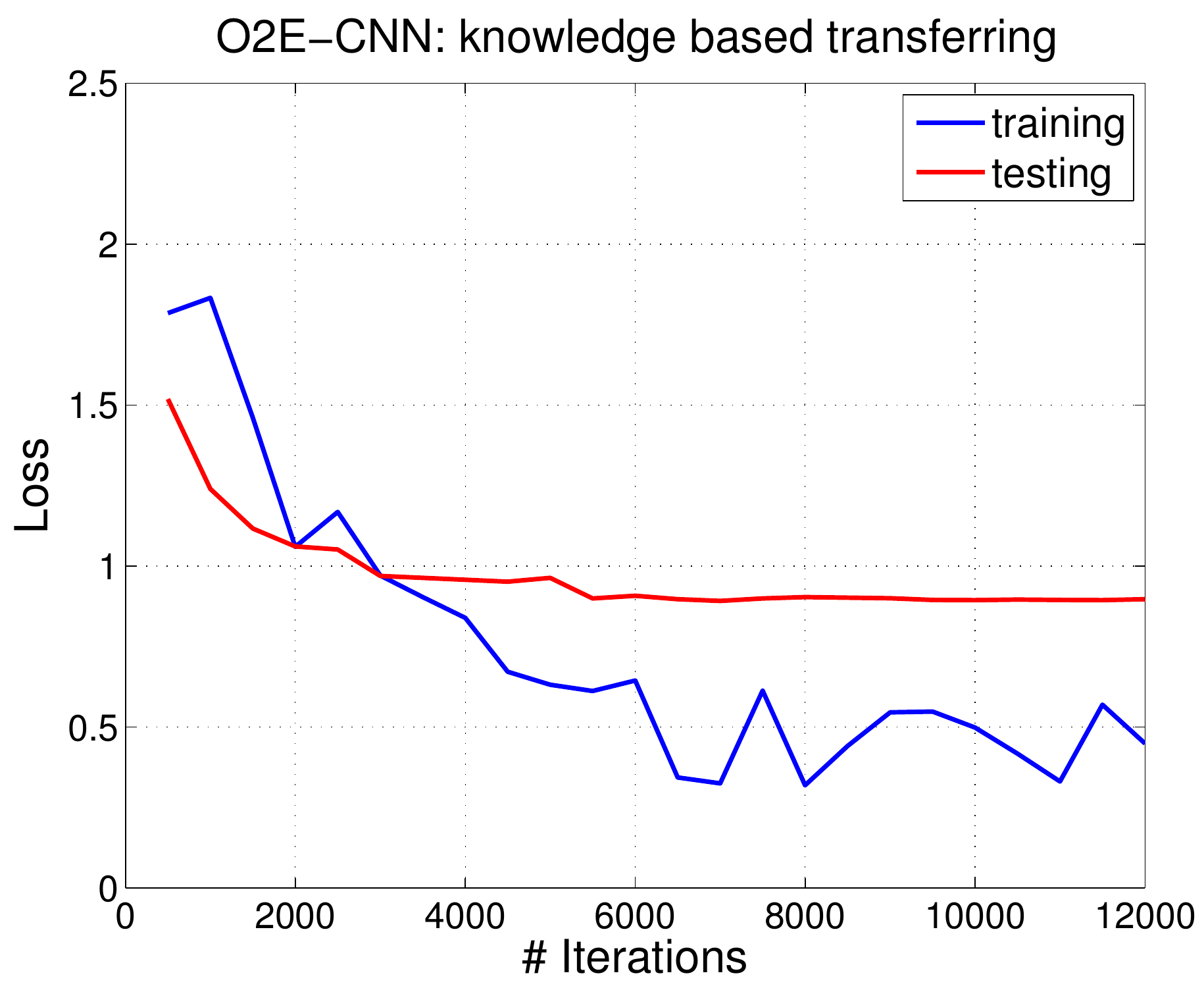}
\includegraphics[width=0.32\textwidth]{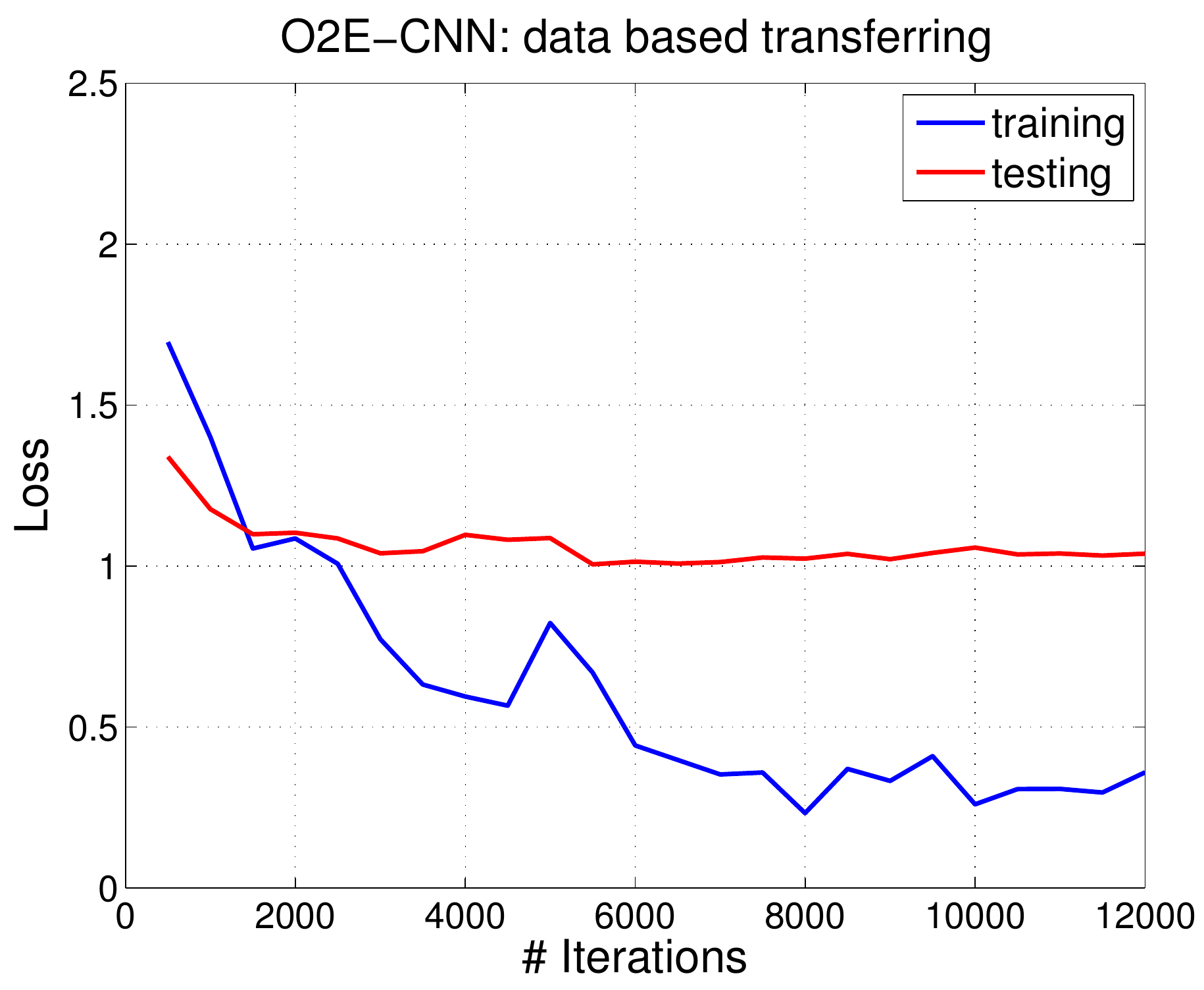} \\
\includegraphics[width=0.32\textwidth]{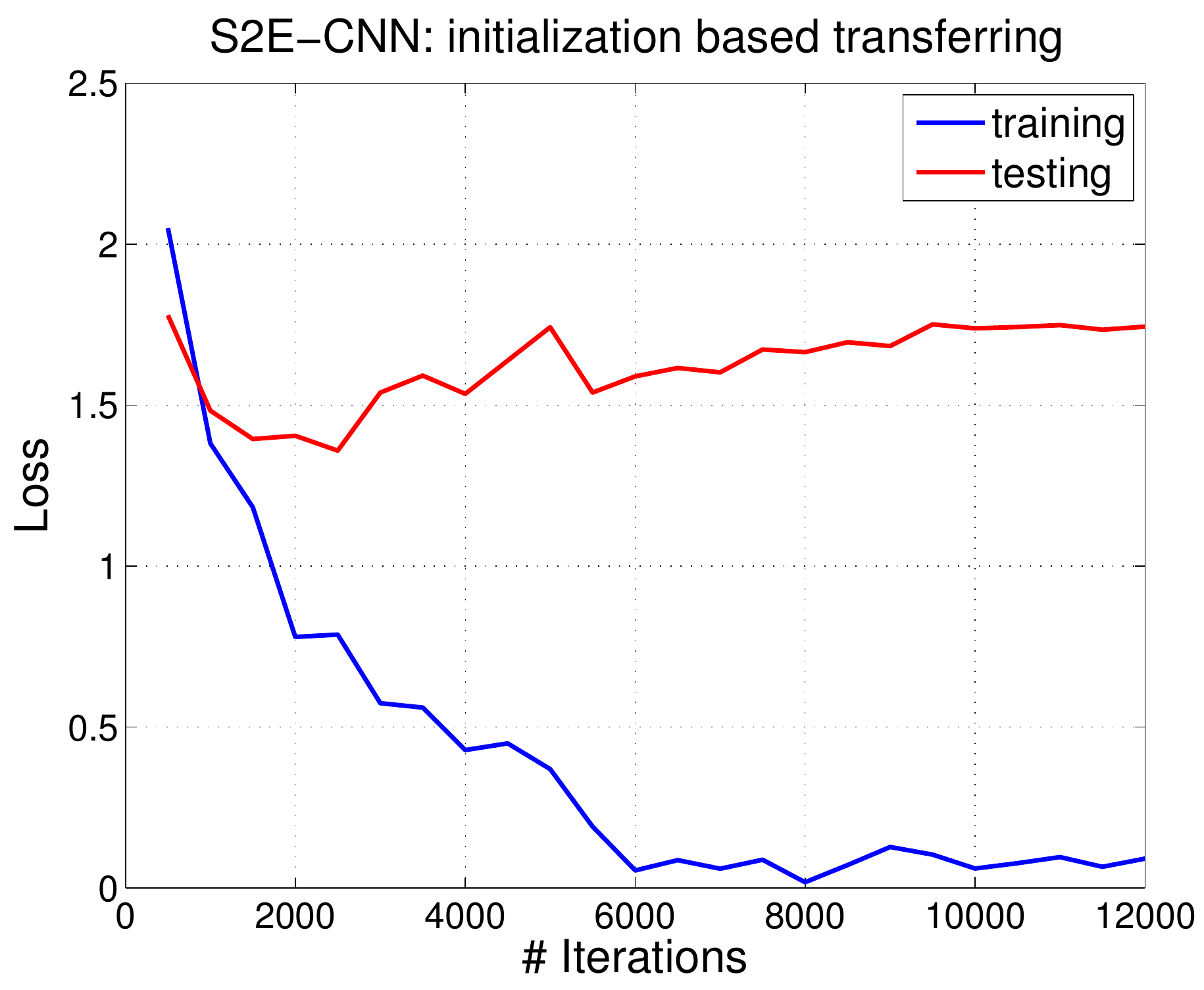}
\includegraphics[width=0.32\textwidth]{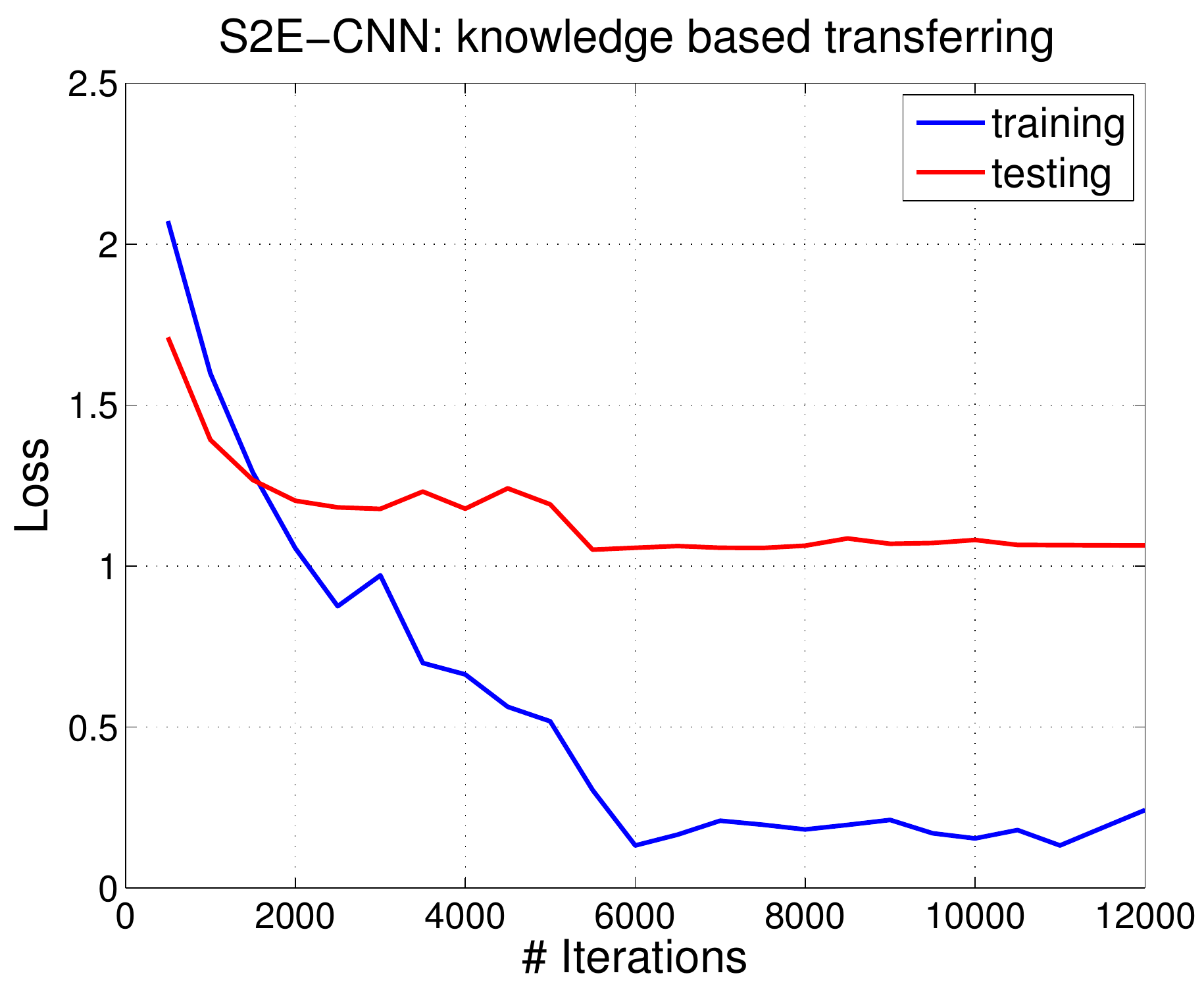}
\includegraphics[width=0.32\textwidth]{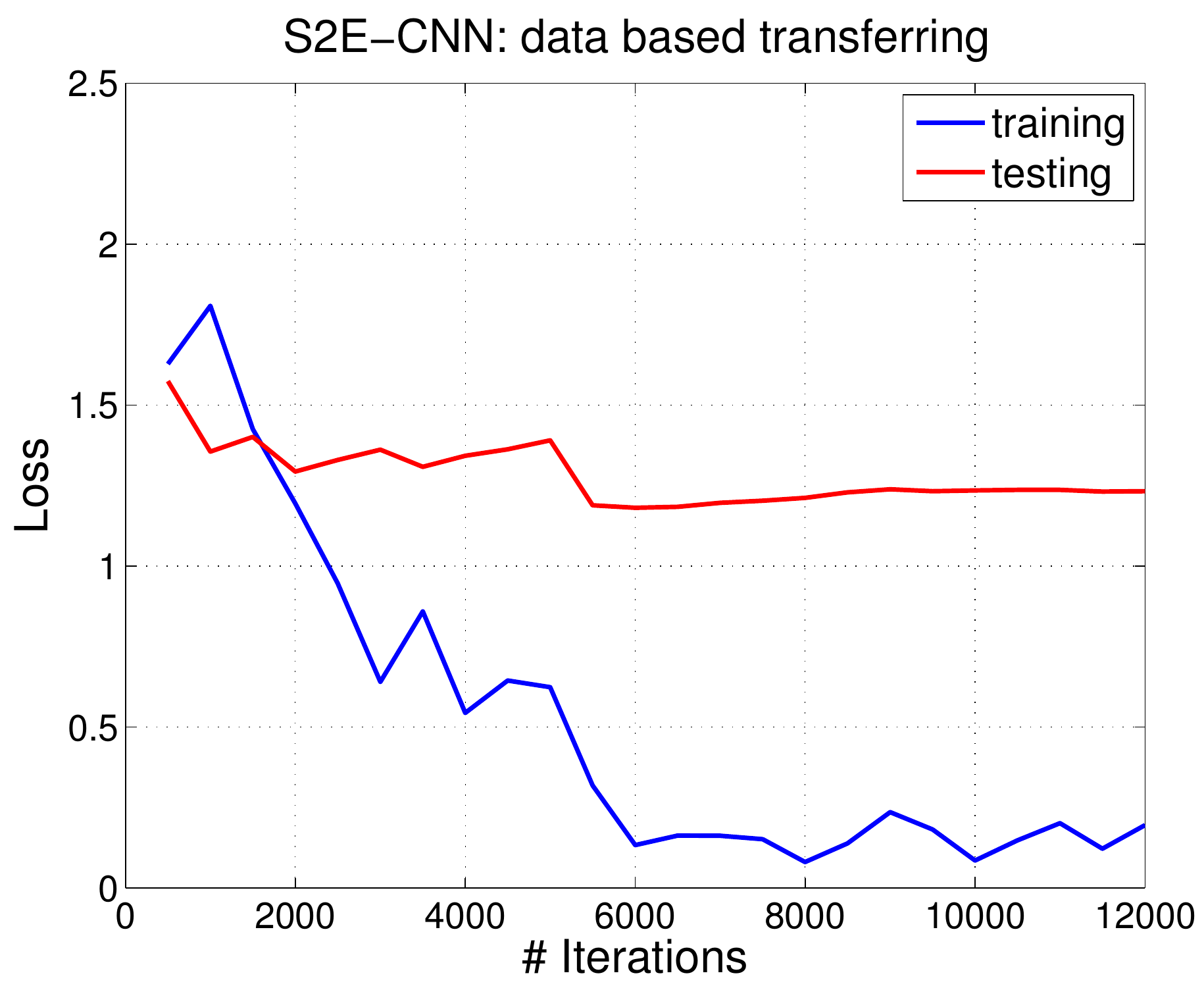}
\caption{Training and testing the loss of O2E-CNN (top row) and S2E-CNN (bottom row) on the ChaLearn Cultural Event Recognition dataset under the {\bf validation setting}. We compare our proposed three transferring techniques. The results indicate that knowledge-based transfer and data-based transfer help to reduce the effect of over-fitting.}
\label{fig:loss}
\end{figure*}

\begin{table}
\caption{Performance of different transferring techniques on the ChaLearn Cultural Event Recognition dataset under the {\bf validation setting}. We compare our proposed three transferring techniques and the data-based transfer achieves the best performance for both O2E-CNN and S2E-CNN.}
\label{tbl:transfer}
\begin{tabular}{|c|c|c|c|}
\hline
Method & O2E-CNNs & S2E-CNNs & OS2E-CNNs \\
\hline \hline
Initialization & 83.9\%& 83.0\%& 85.6\%\\
\hline
Knowledge & 84.8\% & 84.0\%& 86.3\%\\
\hline
Data & 85.5\% & 84.8\%& 87.0\%\\
\hline \hline
Know.+Data & 85.6\%& 85.4\%&87.2\% \\
\hline
ALL & 86.0\%& 85.6\% & 87.2\%\\
\hline
\end{tabular}
\end{table}

After the investigation of cropping strategies and the exploration of parameter settings for multi-task transferring methods, we are ready to study the performance of our different transferring techniques, i.e. the three techniques proposed in Section \ref{sec:transfer}. We test them on the ChaLearn Cultural Event recognition dataset under the validation setting. 

First, we compare the performance of using different pre-trained models: object CNNs pre-trained on the ImageNet dataset and scene CNNs pre-trained on the Places205 dataset. From these results in Table \ref{tbl:transfer}, we observe that deep representations transferred from object CNNs outperform those transferred from scene CNNs. The superior performance of O2E-CNNs may imply that the objects more strongly correlate with events, tallying with the fact that the selected object classes in Figure \ref{fig:selection2} have lower conditional entropy and yield stronger discriminative capacity than the selected scene classes. Furthermore, we fuse the prediction results of O2E-CNNs and S2E-CNNs, enabling further improvements in recognition performance.

Then, we compare the recognition results of the 3 transferring techniques. We see that initialization-based transfer is already effective for fine-tuning event CNNs, and it obtains a performance of $85.6\%$ for OS2E-CNNs. The newly designed knowledge-based transfer and data-based transfer achieve better performance, which indicates that incorporating relevant tasks into the fine-tuning process contributes to improve the generalization ability of the final event models. Data-based transfer is better than knowledge-based transfer but requires additional training images. Furthermore, we fuse the prediction scores of knowledge-based transfer and data-based transfer, boosting the recognition performance a bit further.

Finally, we study the learning procedure in more detail and try to figure out the advantages of incorporating relevant tasks into the fine-tuning pipeline. Specifically, we plot the training and testing loss of three transferring techniques in Figure \ref{fig:loss}. First, we notice that there exist a gap between the training and test loss for all transferring techniques. This indicates over-fitting is still a serious problem for fine-tuning CNNs on a small dataset, even though we initialize those CNNs with pre-trained models. Second, we see that the effect of over-fitting is more severe in the scenario of initialization-based transfer than other scenarios. As the iteration number increases, the test loss stops decreasing and even increases by around 0.3. On the other hand, for knowledge-based and data-based transfer, the extra tasks are helpful to reduce the degree of over-fitting throughout. In summary, from the observation of training and testing loss during fine-tuning, we can see that knowledge-based and data-based techniques are indeed capable of reducing the over-fitting problem and effectively improving the generalization ability.

\subsection{Challenge results}

\begin{table}
\caption{Performance of different transferring techniques on the ChaLearn Cultural Event Recognition dataset under {\bf challenging setting}. Our method outperforms these winners of the ICCV ChaLearn Looking at People (LAP) challenge.}
\label{tbl:challenge}
\resizebox{\linewidth}{!}{
\begin{tabular}{|c|c|c|c|}
\hline
Method & Networks & Explicit Classifiers & Performance \\
\hline \hline
CAS & 4 & LDA+LR & 85.4\\
\hline
FV & 5 & SPM+FV+LR & 85.1\\
\hline
MMLAB & 4 &  FV+SVM &  84.7\\
\hline
CVL\_ETHZ & 2 & LDA+$k$-NN & 79.8 \\
\hline
\hline
Initialization & 2 & none& 85.9 \\
\hline
Knowledge & 2 & none& 86.2 \\
\hline
Data & 2 & none& 86.9 \\
\hline
\hline
Data+Know. & 4 & none & 87.0 \\
\hline
All & 6 & none & {\bf 87.1}\\
\hline
\end{tabular}
}
\end{table}

After the investigation of our method on the ChaLearn Cultural Event Recognition under the {\bf validation setting}, we report the experimental results on this dataset under the {\bf challenging setting}. It is worth noting that we can not access the labels of testing images and the parameter settings are determined according to the study under the validation setting. 

The numerical results are reported in Table \ref{tbl:challenge}. We compare the performance of the proposed method with the winners of the ICCV ChaLearn Looking at People (LAP) challenge \citep{Wei15,Liu15,WangWGQ15,Rothe15}. From these results, we see that the performance of initialization-based transfer achieves a performance of 85.9\%, which outperforms all the winners. This result may be ascribed to the better network structure and the proposed multi-ratio and multi-scale cropping strategy. We also notice that the newly designed multi-task transferring techniques obtain a better performance, and the data-based transfer method gets the best performance of 86.9\% among the three transferring techniques. Finally, we combine the prediction results of different transferring techniques, which yields the best performance of 87.1\% on the test data of the ChaLearn Cultural Event Recognition dataset.

\subsection{Evaluation on other event datasets}

\begin{table}
\caption{Event recognition performance on the Web Image Dataset for Event Recognition (WIDER). We test our proposed transferring methods and compare with the state-of-the-art performance. }
\label{tbl:wider}
\resizebox{\linewidth}{!}{
\begin{tabular}{|c|c|c|}
\hline
Method & Performance \\
\hline \hline
Baseline CNN \citep{XiongZLT15} & 39.7\% \\
\hline
Deep Channel Fusion \citep{XiongZLT15}   & 42.4\% \\
\hline
\hline
Initialization & 50.8\% \\
\hline
Knowledge & 52.0\% \\
\hline
Data & 52.6\% \\
\hline
\hline
Data+Know. & \textbf{53.0\%} \\
\hline
All  & 52.8\%  \\
\hline
\end{tabular}
}
\end{table}

\begin{table}
\caption{Event recognition on the UIUC Sports Event dataset. We test our proposed transferring methods and compare with the state-of-the-art performance. }
\label{tbl:event8}
\resizebox{\linewidth}{!}{
\begin{tabular}{|c|c|}
\hline
Method & Accuracy\\
\hline \hline
Couple LDA \citep{LiF07} & 73.4\% \\
\hline
ImageNet CNN Feature \citep{ZhouLXTO14} & 94.4\% \\
\hline 
Places CNN Feature \citep{ZhouLXTO14} & 94.1\% \\
\hline
GoogLeNet GAP \citep{ZhouKLOT15} & 95.0\% \\
\hline 
\hline
Initialization & 96.9\% \\
\hline
Knowledge & \textbf{98.8\%} \\
\hline
Data & 98.0\% \\
\hline
\hline
Data+Know. & 98.4\% \\
\hline
All  & 98.2\%  \\
\hline
\end{tabular}
}
\end{table}

Having tested our method on the ChaLearn Cultural Event Recognition dataset under both the validation and challenge settings, we present the experimental results on the other event recognition datasets. Specifically, we perform experiments on the Web Image Dataset for Event Recognition (WIDER) \citep{XiongZLT15} and the UIUC Event8 dataset \citep{LiF07}.  We fix the parameter settings used for the ChaLearn Cultural Event Recognition dataset (i.e. no specific tuning).

First, we report the numerical results on WIDER in Table \ref{tbl:wider}. We see that our newly proposed transferring methods outperform initialization-based transfer, in keeping with our findings on the ChaLearn Cultural Event Recognition dataset. Knowledge-based transfer and data-based transfer improve the performance of initialization-based transfer by around $1\%$ and $2\%$, respectively. Combining the prediction results of knowledge-based and data-based transferring further boosts the recognition performance. We compare the performance of our method with two other approaches: (1) baseline CNN models and (2) deep channel fusion \citep{XiongZLT15}, which obtained the state-of-the-art performance on this dataset. Our transferring method significantly outperforms these two methods by a large margin of around $11\%$.

Second, the results on the UIUC Event8 dataset are summarized in Table \ref{tbl:event8}. This dataset is relatively small and the state-of-the-art performance is very high (around 95\%).  We notice that our baseline of initialization-based transfer achieves a performance of $96.9\%$, and our novel transferring techniques are still able to boost the performance to $98.0\%$. This superior performance of knowledge-based and data-based transfer proves the effectiveness of incorporating auxiliary tasks in the fine-tuning. The performance of data-based transfer is a bit lower than that of knowledge-based transfer. This can be ascribed to the smaller size of the UIUC Event8 dataset, which requires a small number of iterations for convergence. So, the fine-tuning is not able to fully exploit the extra ImageNet and Places205 images before convergence. We compare to the baseline of Couple LDA \citep{LiF07} and other recently proposed deep learning methods \citep{ZhouLXTO14,ZhouKLOT15}. Our proposed transferring methods outperform these previous approaches and obtain the state-of-the-art performance of $98.8\%$ on this dataset.

\subsection{Visualization of recognition examples}
\begin{figure*}[t]
\includegraphics[width=\textwidth]{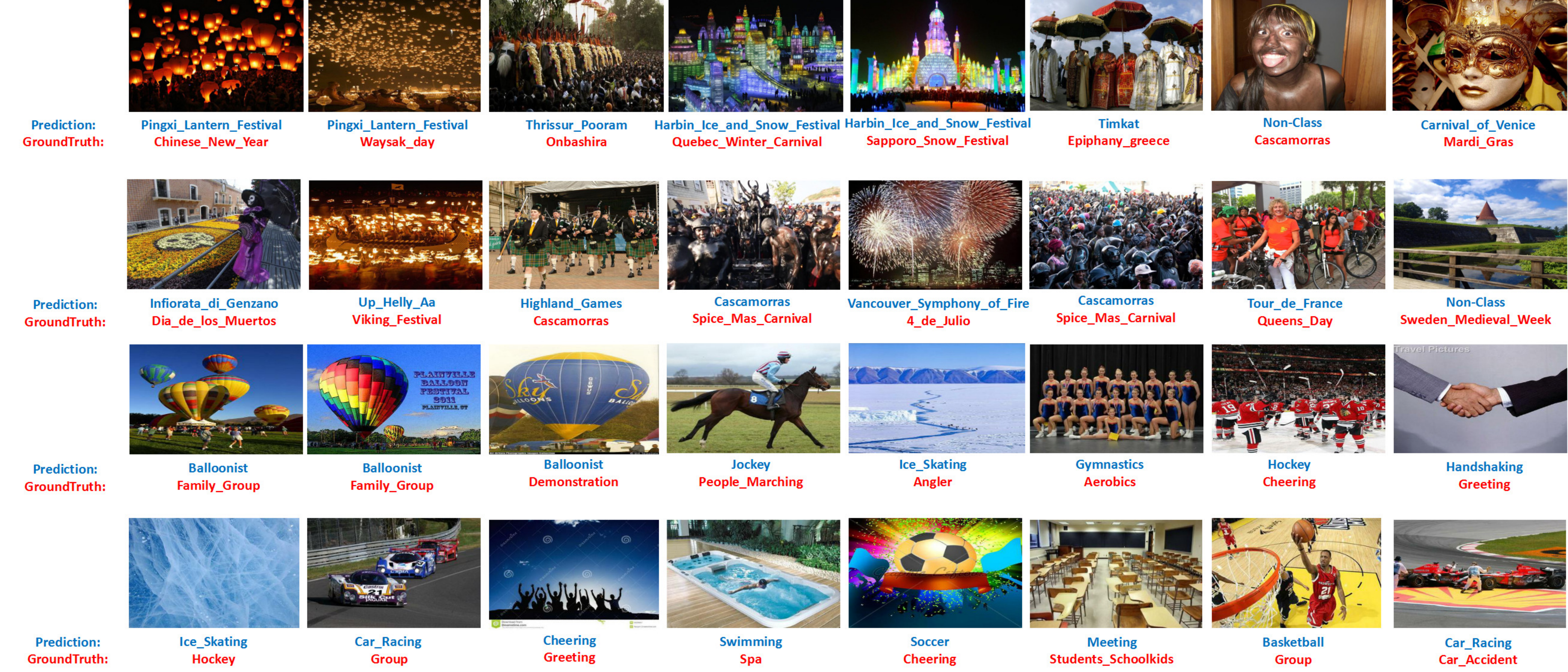}
\caption{Examples of images that our method fails on. We show several failure cases from the ChaLearn Cultural Event Recognition dataset at the top 2 rows and from the Web Image Dataset for Event Recognition (WIDER) at the bottom 2 rows. We notice that sometimes the ground truth labels contain noise and our prediction results seem more reasonable. }
\label{fig:result_example}
 \end{figure*}
 
Several wrong event recognition examples are given in Figure \ref{fig:result_example}. In these cases, our method produces a wrong label with high confidence. In the top 2 rows, we show some failure cases from the ChaLearn Cultural Event Recognition dataset. From these samples, we see that the event class \texttt{chinese new year} may be easily confused with the event class \texttt{pingxi lattern festival}, that the event \texttt{harbin ice and snow festival} comes close in appearance to the events \texttt{sapporo snow festival} and \texttt{quebec winter carnival}, that the event class \texttt{carnival of venice} looks like \texttt{mardi gras}, and so on. In the bottom 2 rows, we give some failure cases for the Web Image Dataset for Event Recognition (WIDER).  We see that our method may confuse the class \texttt{balloonist} with the class \texttt{family group}, the class  \texttt{jockey} with the class of \texttt{people marching}, the class \texttt{gymnastics} with the class \texttt{aerobics}, and so on. Also, we notice that sometimes the ground truth labels contain noise and our prediction results seem more reasonable.


\section{Conclusions}
\label{sec:conclusion}

We have presented a deep architecture, coined OS2E-CNN, for event recognition in still images. It transfers deep representations from object and scene models to the event recognition domain. Objects, scenes, and events are indeed semantically related. We empirically studied the relation between objects, scenes, and event classes. It appears that the likelihood of object and scene classes matters for event understanding in still images. Yet, not all object and scene classes strongly correlate with the event classes, and we designed an effective method to select a subset of discriminative and diverse object and scene classes. To adapt these deep learned representations of object and scene models, we developed three transferring techniques: (1) initialization-based transfer, (2) knowledge-based transfer, and (3) data-based transfer. The latter two transferring techniques exploit multi-task learning frameworks to incorporate the extra knowledge from other networks or extra data from public datasets into the fine-tuning of event models. It turns out that these new transferring methods are effective to reduce over-fitting and to improve the generalization ability. Our method achieves the state-of-the-art performance and outperforms competing approaches on three public benchmarks.  

In the future, we may consider incorporating more visual semantic cues such as human pose, garments, etc. into a unified framework for event recognition from still images. The concept of event is a higher-level concept than other semantic ones such as objects and scenes, and we investigate into a new recognition framework to exploit the hierarchical structure among the task of object recognition, scene recognition, and event recognition.

\small
\bibliographystyle{spbasic}
\bibliography{event}

\end{document}